\documentclass[10pt]{article} 
\usepackage[accepted]{tmlr}

\usepackage{amsmath,amsfonts,bm}

\def\eqref#1{equation~\ref{#1}}

\def\1{\bm{1}}

\def\va{{\bm{a}}}
\def\vb{{\bm{b}}}
\def\vc{{\bm{c}}}

\def\vh{{\bm{h}}}

\def\vo{{\bm{o}}}

\def\vr{{\bm{r}}}

\def\vv{{\bm{v}}}
\def\vw{{\bm{w}}}
\def\vx{{\bm{x}}}

\def\vz{{\bm{z}}}

\DeclareMathAlphabet{\mathsfit}{\encodingdefault}{\sfdefault}{m}{sl}
\SetMathAlphabet{\mathsfit}{bold}{\encodingdefault}{\sfdefault}{bx}{n}

\newcommand{\R}{\mathbb{R}}

\usepackage[hidelinks]{hyperref}
\usepackage{url}
\usepackage{xspace}
\usepackage{bbm}
\usepackage{import}
\usepackage{marginnote}
\usepackage{mathabx}
\usepackage{booktabs}
\usepackage{neuralnetwork}
\usepackage{caption}
\usepackage{subcaption}
\usepackage{amsmath}
\usepackage{amssymb}
\usepackage{amsthm}
\usepackage{mathtools}
\usepackage{graphicx}

\usepackage{wrapfig}
\usepackage{algorithm}
\usepackage{algpseudocode}
\usepackage{siunitx}   
\usepackage{listofitems} 
\usetikzlibrary{arrows.meta} 
\usepackage[outline]{contour} 
\let\underbrace\LaTeXunderbrace

\newtheorem{theorem}{Theorem}
\newtheorem{proposition}{Proposition}

\theoremstyle{definition}
\newtheorem{definition}{Definition}
\newtheorem{remark}{Remark}
\newtheorem{assumption}{Assumption}

\makeatletter
\newcommand{\subalign}[1]{%
  \vcenter{%
    \Let@ \restore@math@cr \default@tag
    \baselineskip\fontdimen10 \scriptfont\tw@
    \advance\baselineskip\fontdimen12 \scriptfont\tw@
    \lineskip\thr@@\fontdimen8 \scriptfont\thr@@
    \lineskiplimit\lineskip
    \ialign{\hfil$\m@th\scriptstyle##$&$\m@th\scriptstyle{}##$\hfil\crcr
      #1\crcr
    }%
  }%
}
\makeatother
\newcommand{\idx}[1]{{\scriptscriptstyle [#1]}}

\newcommand{\ind}{\mathbbm{1}}

\newcommand{\method}[1]{\text{\small \texttt{.#1}}}

\title{
    A Rigorous Study Of The Deep Taylor Decomposition
}

\author{%
    \name Leon Sixt \email leon.sixt@fu-berlin.de \\
    \addr Department of Computer Science\\
    Freie Universit\"at Berlin
    \AND
    \name Tim Landgraf \email tim.landgraf@fu-berlin.de \\
    \addr Department of Computer Science\\
    Freie Universit\"at Berlin
}

\def\month{11}  
\def\year{2022} 
\def\openreview{\url{
https://openreview.net/forum?id=Y4mgmw9OgV
}} 

\begin{document}

\maketitle

\def\LRPa{LRP$_{\alpha1\beta0}$\xspace}
\def\LRPab{LRP$_{\alpha2\beta1}$\xspace}
\def\LRPe{LRP$_{\varepsilon}$\xspace}
\def\LRPO{LRP$_{0}$\xspace}

\begin{abstract}

Saliency methods attempt to explain deep neural networks by highlighting the
most salient features of a sample.
Some widely used methods are based on a theoretical framework called Deep Taylor Decomposition (DTD), which formalizes
the recursive application of the Taylor Theorem to
the network's layers.
However, recent work has found these methods to be independent of the network's deeper layers and appear to respond only to lower-level image structure.
Here, we investigate
the DTD theory
to better understand
this perplexing behavior
and found that
the Deep Taylor Decomposition is equivalent to the basic gradient$\times$input method when
the Taylor root points (an important parameter of the algorithm chosen by the user) are locally constant.
If the root points are locally input-dependent, then one can justify any explanation.
In this case,
the theory is under-constrained.
In an empirical evaluation, we find that DTD roots do
not lie in the same linear regions as the input --
contrary to a fundamental assumption of the Taylor theorem.
The theoretical foundations of DTD
were cited as a source of reliability for the
explanations. However, our findings urge caution in making such claims.

\end{abstract}

\section{Introduction}

Post-hoc explanations are popular for explaining Machine Learning models as they do not require changing the model's architecture or training procedure.
In particular, feature attribution methods are widely used. They assign a
saliency score to each input dimension, reflecting their relevance for the model's output. For images, the saliency scores can be visualized as heatmaps (see Figure \ref{fig:sanity_checks}).

Evaluating post-hoc explanations is challenging because it is inherently circular:
As we do not understand the internal workings of the model, which we are trying
to explain, we cannot judge the quality of the explanation.
The situation is further complicated as many methods simplify the model's
complexity to render explanations accessible to the human eye.
For example, most methods focus on the local neighborhood of an input sample, and rely on assumptions such as linearity (e.g. gradient-based methods) or independence of the input features (e.g. approximation of Shapley values \citep{vstrumbelj2011general,vstrumbelj2014explaining,lundberg2017unified,kumar2020problems}).

These factors and the complexity of deep neural networks make it difficult to assess whether an explanation is correct or not. We can not disentangle failures of the explanation method and unexpected behavior of the model.
While it is acceptable for methods to introduce simplifications or rely on assumptions, their existence, purpose, and violation should be made transparent.
In the best case, a method would be based on a solid theoretical foundation providing guarantees regarding an explanation's correctness.

Such a theoretical foundation is the
\emph{Deep Taylor Decomposition} (DTD, \cite{montavon2017dtd}).
DTD recursively applies the Taylor Theorem to the network's layers,
and backpropagates modified gradients to the input, thereby computing the input's relevance.
It was used as theoretical foundation of LRP
\citep{bach_pixel-wise_2015}, a poplar method to explain image models.
LRP was repeatedly advertised as a sound and reliable explanation
technique \citep{montavon2019overview,samek_explaining_2021,holzinger2022explainable}.
For example, \cite{holzinger2022explainable} stated:
    \emph{``The main advantages of LRP are its
    high computational efficiency [...], its
    theoretical underpinning making it a trustworthy and robust explanation method
    [...], and its long tradition
    and high popularity [...].''
}

However,
\cite{sixt2020wel} has shown that
certain LRP and DTD backpropagation rules
create explanations partially independent of the model's parameters:
the explanation will remain the same
even if the last layer's parameters are randomized.
The theoretical analysis in \citep{sixt2020wel}
revealed that the propagation matrices,
which correspond to the layers' Jacobian matrices,
are all positive and their product converges to a rank-1 matrix quickly.
To obtain the saliency map, the result is usually normalized,
and thereby even the last single degree of freedom is lost.
Thus, the explanation does not change when explaining a different class or the parameters of the deeper layers is changed.

This perplexing behavior questions the consistency of DTD directly.
While \cite{sixt2020wel} described the convergence to the rank-1 matrix in detail,
the failure was not related to DTD's theory such as the choice of root points\footnote{%
        Following \cite{montavon2017dtd},
        we name the points used for the Taylor Theorem
        \emph{root points}. For example,
        for a function $f: \R \to \R$,
        the first-order Taylor approximation is
        $f(x) \approx f(\tilde x) + f'(\tilde x)(x - \tilde x)$,
        where $\tilde x$ is the root point.}
and the
recursive application of the Taylor Theorem.
Here, we fill this gap: \emph{%
Can we identify flaws in DTD's theory that would explain the perplexing behavior of ignoring the network's parameters? Does DTD provide transparency regarding its assumptions and guarantees about the explanation's correctness? }

Before we approach these questions, we summarize the relevant background of the Deep Taylor Decomposition in Section \ref{sec:background}.
For completeness, we start with the well-known Taylor theorem and then discuss how the theorem connects to DTD's relevances. We then continue with stating the recursive application of the Taylor Theorem formally
and recapitulate the so-called \emph{train-free DTD} approximation, which allows to compute layer-wise relevances efficiently.

In section \ref{sec:theoretic_analysis},
we present our theoretical analysis of DTD.
In particular, we contribute:
\textbf{(C1)}
a proof that the root points
 must be contained in the same linear region as the input;
\textbf{(C2)} we generalize a previous observation about \LRPO \citep{avanti2016blackbox}:
if the layers' root points are chosen locally constant w.r.t
the layers' input, then DTD's relevances take a similar form as input$\times$gradient;
\textbf{(C3)}
DTD is under-constrained:
if the root points depend on the layers' input, then the Deep Taylor Decomposition can be used to create any arbitrary explanation;
\textbf{(C4)} we also find that DTD cannot be extended easily to analytic activation functions
(e.g. Softplus), without introducing complex higher-order derivatives with the same order as
the number of network layers.

In an empirical evaluation (Section \ref{sec:experiments}),
we applied the theoretical insights from the previous section
and studied the \emph{train-free DTD} approximation in several experiments:
\textbf{(C5)} The train-free DTD does not enforce
the root points to be located in the valid
local linear region of the network;
\textbf{(C6)} We also validated this empirically using a small multilayered perceptron,
where we found a substantial number of samples having roots located outside the valid local linear region;
\textbf{(C7)} Additionally, we include a reproducibility study of \citep{arras2022clevr}
that claimed that DTD's explanations would not suffer from
the problems reported in \cite{sixt2020wel}.
This reproducibility study also highlights DTD's black-box
character and how difficult it is to evaluate
explanation quality empirically.

Given the theoretical and empirical evidence, we conclude that DTD obscures its
simplifications and violates its own assumptions. DTD is underconstrained and
even allows justifying virtually any explanation.

\section{Related Work}

The theoretical analysis of explanation methods is a small
research area.
PatternAttribution \citep{kindermans_learning_2018} investigated the insensitiveness of DTD rules to input noise and then proposed a way to learn the root points from data.
Other lines of work are the manipulation of saliency maps
\citep{dombrowski2019explanations,viering2019manipulate,wang-etal-2020-gradient},
the runtime-complexity of explanation methods
\citep{waeldchen2021computational},
or
the explanations methods with provable guarantees  \citep{chen2018lshapley}.

Previous works have analyzed the theoretical properties of various saliency methods.
For example, the insenstivity of Guided-Backprop \citep{springenberg2014striving} was analyzed in \citep{nie_theoretical_2018}, and
\citep{lundstrom2022int_grad}
found flaws in the theoretical motivation of integrated gradients \cite{sundararajan2017axiomatic}.
\citep{kumar2020problems}
discussed the 
issues from an independence assumption
between input variables, often introduced 
in sampling algorithms \citep{vstrumbelj2011general,vstrumbelj2014explaining,lundberg2017unified}.
In  \citep{shah2021grads}, it was empirically analyzed and proven
for a specific dataset that the Gradient's magnitude will not correspond to relevant features.
Our work also analyzes the theoretical properties of saliency methods but differs from previous works as it focuses 
on the DTD and LRP methods.

Although different review articles
\citep{montavon2018review,montavon2019overview,samek_explaining_2021,samek2021conv_rnn}
and extensions of LRP and Deep Taylor
\citep{binder_layer-wise_2016,kohlbrenner2020towards,hui_batchnorm_2019,ali2022transformers}
have been published, none discussed the theoretical issues brought forward in our
manuscript.

\section{Background}
\label{sec:background}

In this section, we
provide the necessary background on Deep Taylor Decomposition to understand the theoretical
analysis in Section \ref{sec:theoretic_analysis}.
We mainly reproduce the derivations given in
\citep{montavon2017dtd,montavon2018review}.
If we comment on the derivations, we do this in the \emph{Remark} sections.

\subsection{Taylor Theorem for multivariate functions}

\def\valpha{{\bm{\alpha}}}
Taylor Theorem for multivariate functions can be concisely stated using multi-index notation.
A multi-index $\bm{\alpha} \in \mathbb{N}_0^k$ is a vector of non-negative integers ($\bm{\alpha} = [\alpha_1, \ldots, \alpha_k]$).
The following operations are defined as:
$\valpha! = \alpha_1!\alpha_2!\ldots\alpha_k!$, $|\bm{\alpha}| = \sum_{i=1}^k \alpha_i$,
$\vx^\valpha = x_1^{\alpha_1} x_2^{\alpha_2} \ldots x_k^{\alpha_k}$, and
$\partial^\valpha f =
\partial^{|\valpha|} f /
(\partial^{\alpha_1} x_1
\partial^{\alpha_2} x_2
\ldots
\partial^{\alpha_k} x_k
)$, where $\vx \in R^k$ and $f: \R^{k} \to \R$.
The following theorem is adapted from \citet[Theorem 2.68]{folland_advanced_2002}:
\begin{theorem}[Multivariate Taylor Theorem]
Suppose  $f: \R^d \to \R$ is of differentiability class $C^{k}$  on an open convex set $S$.
If $\vx \in S$ and $\tilde \vx \in S$, then:
\vspace{-0.25cm}
\begin{equation}
    f(\vx) = \sum_{|\valpha| \le k }
        \frac
            {\partial^\valpha f(\tilde \vx)}
            {\valpha!}
        (\vx - \tilde \vx)^{\valpha}
    + g_{ k}(\vx, \tilde \vx),
\end{equation}
\vspace{-0.85cm}

where the remainder is given by:
\vspace{-0.20cm}
\begin{equation}
    g_{k}(\vx, \tilde \vx) =
        k \sum_{|\valpha| = k}
        \frac
            {(\vx - \tilde \vx)^\valpha}
            {\valpha!}
        \int_0^1
            (1 - t)^{k-1}
            \Big[
                \partial^\valpha f\big(
                    t \vx +
                    \left(1-t\right) \tilde \vx
                \big)
                - \partial^\valpha f(\vx)
            \Big]
        dt.
\end{equation}
\end{theorem}
\vspace{-0.40cm}
As the Deep Taylor Decomposition focuses on neural networks with ReLU activations,
we will mainly look at the first-order Taylor Theorem:
\begin{equation}
    \label{eq:first_order_taylor_theorem}
    f(\vx) = f(\tilde \vx)
        + \frac
            {\partial f(\vx)}
            {\partial \vx}
          \Big|_{\vx = \tilde \vx}
          \cdot (\vx - \tilde \vx)
\end{equation}
where $\tilde \vx$ is the root point, and $|_{\vx = \tilde \vx } $ denotes
the gradient evaluated at the root point $\tilde \vx$.
The higher order terms are zero due to the local linearity of ReLU networks.
As the Taylor Theorem requires $f \in C^1$ (i.e.,
all partial derivatives $\partial f(\vx) / \partial \vx$
must be continuous in the local neighborhood $S$),
the root point must be within the same linear region as the input.

\begin{definition}[Linear Region]
    A linear region of a function $f: \R^d \to \R$ is the set $N_f(\vx)$ of all points
    $\vx' \in N_f(\vx)$ that (1) have the same gradient at $\vx$: $
    \,\nabla f(\vx) = \nabla f(\vx')
    $, and (2)
    can be reached from $\vx$ without passing through a point $\va$ with a
    different gradient, i.e., $ \nabla f(\va) \neq \nabla f(\vx)$.
\end{definition}
In case of a ReLU network, the approximation error cannot be bounded when
selecting a root point $\tilde \vx \not \in N_f(\vx)$, as that linear region's
gradient might differ substantially.

\subsection{Taylor Theorem and Relevances}
\label{subsec:taylor_theorem_and_relevances}

In the previous section, we have recapitulated the Taylor Theorem.
We now discuss how the Taylor Theorem can be used to compute input relevances.
A common approach explaining deep neural network is to contrast the
network's output with a similar point predicted differently.
A user could then study pairs $(\vx, f(\vx))$ and $(\tilde \vx, f(\tilde \vx))$
and relate the input differences to the output differences.
To guide the user's attention,
it would be desirable to highlight which changes between $\vx$
and $\tilde \vx$ were responsible for the difference in the output. The
first observation in \citep{montavon2017dtd} is that the network's output
differences can be redistributed to the input by using the Taylor Theorem.
If the point $\tilde \vx$ is in the local neighborhood $N_f(\vx)$,
we can use the first-order Taylor Theorem (\eqref{eq:first_order_taylor_theorem}) to
write the difference $f(\vx) - f(\tilde \vx)$ as:
\begin{equation}
    f(\vx) - f(\tilde \vx) =
        \frac
            {\partial f(\vx)}{\partial \vx}\big|_{\vx = \tilde \vx}
            \cdot
            (\vx - \tilde \vx)
\end{equation}
The relevance of the input $R: \R^d \to \R^d$ is then defined to be the
point-wise product of the partial derivatives with the input differences:
\begin{equation}
    R(\vx) = \frac
        {\partial f(\vx)}{\partial \vx}\big|_{\vx = \tilde \vx}
        \odot
        (\vx - \tilde \vx)
\end{equation}
While this would be a simple way to compute the relevances,
the following reasons are given
in \citep{montavon2017dtd,montavon2019overview},
to not directly use the Taylor Theorem on the network output:

\begin{enumerate}
\item
\textbf{Adversarial perturbations} \citep{szegedy2013intriguing}:
    Small input perturbations can lead to a large change in the output.
    Therefore, the difference in the output might be enormous but $|\vx - \tilde \vx|$ tiny and uninterpretable.
\item
\textbf{Finding a root point might be difficult}:
    \emph{``It is also not necessarily solvable due to the possible non-convexity of
    the minimization problem''} \citep{montavon2017dtd}.
\item
 \textbf{Shattered gradients} (\cite{balduzzi_shattered_2017}): \emph{``While the function value $f(x)$ is
    generally accurate, the gradient of the function is
    noisy''} \citep{montavon2019overview}.
\end{enumerate}

\begin{remark}
We want to point out that the more general problem seems to be that the local
linear regions are tiny, or rather the number of linear regions grows
exponentially with the depth of the network
in the worst case
\citep{arora_understanding_2018,xiong_number_2020,montufar_number_2014}.  This
restricts the valid region for the root point to a small neighborhood around the
input.
\end{remark}

\subsection{Deep Taylor: Recursive Application of Taylor Theorem}
\label{subsec:dtd_recursive}

The main idea of \citep{montavon2017dtd} is to recursively apply the Taylor Theorem
to each network layer. Before we present this in detail, we first we need to
clarify the notation of an n-layered ReLU network shortly:
\begin{definition}[ReLU network]
    An $n$-layered ReLU network $f: \R^{d_1} \to \R_{\ge0}^{d_{n+1}}$ is the composition
    of $n$ functions $f = f_n \circ \ldots \circ f_1$, where
    each function
    $f_l: \R^{d_l} \to \R_{\ge0}^{d_{l+1}}$
    has the form
    $f_l(\va_l) = \left[W_l \va_l\right]^+$,
    and where $[.]^+$ is the ReLU activation.
\end{definition}

Instead of directly calculating the relevance
of the input as done in the previous section,
we can apply Taylor Theorem to the final network layer
and then apply the Taylor Theorem again to the resulting
relevance. By recursively applying the Taylor Theorem per individual layer,
we can calculate the relevance of the input.
As the base case of the recursive application, the
relevance of the network output is set to the value of the explained logit
$f_\idx{\xi}(\vx)=\va_{{n+1}_\idx{\xi}}$:
\begin{equation}
    \label{eq:dtd_recursive_base_case}
    R^{n+1}(\va_{n+1}) = \va_{n+1_\idx{\xi}},
\end{equation}
where $R^{n+1}$ denotes the relevance of the $n+1$-th network activation. We decided to
use superscripts for the relevance functions
as their individual dimensions are often index as in $R^{n+1}_\idx{j}$.
Suppose that we already know the relevance function $R^{l+1}(\va_{l+1}) \in \R^{d_{l+1}}$ for the layer $l\!+\!1$.
We can then calculate the relevance of $\va_l$ (the input to layer $l$)
to the $j$-th coordinate of $R^{l+1}(\va_{l+1})$:
\begin{equation}
    R^{l+1}_\idx{j}(\va_{l+1}) = R^{l+1}_\idx{j}\left(f_l(\tilde \va_{l})\right)
        +
        \frac
            {
                \partial R^{l+1}_\idx{j}\left(
                    f_l(\va_{l})
                \right)
            }
            {\partial \va_{l}}
            \Bigg|_{\va_{l} =\tilde \va_l(\va_l)}
            \cdot
            (\va_l - \tilde \va_{l}(\va_l)),
\end{equation}
where we used $\va_{l+1} = f_l(\va_l)$. The root point is selected in dependency of the layers' input $\va_l$, i.e., it is a function $
 \tilde \va_l: \R^{d_l} \to \R^{d_{l}}
$.
The total relevance of the input to layer $l$ is given by the sum over all $d_{l+1}$ hidden neurons.

\begin{definition}[Recursive Taylor]
\label{def:dtd_recursive_taylor}
Given a function $f: \R^{d_1} \to \R^{d_{n+1}}$, which can be written as a composition of $n$ functions
$f = f_1 \circ \ldots \circ f_{n}$ with $f_l: \R^{d_l} \to \R^{d_{l+1}}$,
the input to each function $f_l$ are denoted by $\va_l$ and $\va_{n+1}$ specifics
$f$'s output.
Additionally, a root point function $\tilde \va_l: \R^{d_l} \to \R^{d_l}$
is defined for each layer, which must only return admissible values $\tilde
\va_l(\va_l) \in N_{R_l}(\vx)$ and $\tilde \va_l(\va_l) \neq \va_l$.
Then, the
base case is given by $R^{n+1}(\va_{n+1}) = \va_{n+1_\idx{\xi}}$
and the relevance function $R^l: \R^{d_l} \to \R^{d_l}$ of layer $l \neq n$ is
recursively defined by:
\begin{equation}
\label{eq:dtd_recursive_taylor_step}
R^{l}(\va_l) = \sum_{j = 1}^{d_{l+1}}
    \left(
        \frac
            {
                \partial R^{l+1}_\idx{j}\left(
                    f_{l}(\va_{l})
                \right)
            }
            {\partial \va_{l}}
            \Bigg|_{\va_{l} =\tilde \va_l^{(j)}(\va_l)}
            \odot
            \left(\va_l - \tilde \va_{l}^{(j)}(\va_l)\right)
    \right)
\end{equation}
\end{definition}
The above definition corresponds to equation 6 in \citep{montavon2017dtd}.
Except for the root point selection, which will be discussed in the following sections,
definition \ref{def:dtd_recursive_taylor} contains all information to implement
the recursive decomposition using an automatic differentiation library.
An exemplary pseudo-code can be found in Algorithm \ref{alg:dtd_recursive_taylor}.
Before continuing with the approximations of the Deep Taylor Decomposition, we want to make a few remarks:

\begin{remark}[No Axiomatic Motivation]
Only some vague arguments are provided to motivate the recursive application of the Taylor
Theorem:
\vspace{-0.25cm}
\begin{quotation}
    \emph{
        The deep Taylor decomposition method is inspired by the
        divide-and-conquer paradigm, and exploits the property that the function
        learned by a deep network is decomposed into a set of simpler
        subfunctions, either enforced structurally by the neural network
        connectivity, or occurring as a result of training.
    } -- \citet[Section 3]{montavon2017dtd}
\end{quotation}
\vspace{-0.25cm}
In contrast, Shapely values \citep{shapley1951} are motivated by four axiomatic properties,
which are uniquely fulfilled by the Shapely values.
A comparable set of axioms with uniqueness result does not exist for the Deep
Taylor Decomposition.
\end{remark}

\subsection{Deep Taylor Decomposition For A One-Layer Network and DTD's rules}
\label{subsec:dtd-single-layer}

In the previous section, we introduced the recursive application of the Taylor Theorem.
For a concrete example,
we will now discuss how DTD is applied to a one-layered network. It will also
explain the propagation rules of DTD.
This subsection corresponds to Section 4 in \cite{montavon2017dtd}, and we will refer to the
corresponding equations with the notation (DTD eq. 11).

The one-layered network consists of a linear layer with a ReLU activation followed by a sum-pooling
layer, i.e.
$f: \R^d \to \R$, $f(\vx) = \sum_j [W \vx + \vb]_j^+$,
where $[ . ]^+$ is the ReLU activation.
We will denote the output of the ReLU layer as
$\vh(\vx) = [W \vx + \vb]^+$,
and the sum-pooling layer as $y(\vh) = \sum_j \vh_j$.
For this subsection, we will denote the relevance function with $R^\vx$, $R^\vh$, and $R^y$
for the input, hidden, and output layer, respectively.

The relevance of the final layer is simply given by the network's output (\eqref{eq:dtd_recursive_base_case}; DTD eq. 8):
\begin{equation}
    R^y(\vh(\vx)) = y(\vh(\vx)).
\end{equation}
DTD suggests to select a root point $\tilde \vh$ such that $y(\tilde \vh) = 0$.
The advantage of $y(\tilde \vh) = 0$ is that the network's output $y(\vh)$ is absorbed to
the first-order term (i.e., $f(\tilde \vx) = 0$ in \eqref{eq:first_order_taylor_theorem}).
We then have $y(\vh) = \frac{\partial R^y(\tilde \vh)}{\partial \tilde \vh}
\cdot (\vh - \tilde \vh)$ such that the network output is fully redistributed
to the hidden layer's relevance.
Additionally, the root point should be a valid input to the layer.
As $y(\vh)$'s input comes from the ReLU layer $\vh(\vx)$, it is positive
and only $\tilde \vh = 0$ solves $\sum_j \tilde \vh_j = 0$.
The derivative
$\partial R_\vh^f(\tilde \vh) / \partial \tilde \vh =
\partial y(\tilde \vh) / \partial \tilde \vh = 1$
 and therefore, we can use \eqref{eq:dtd_recursive_taylor_step} to write the relevance of the ReLU layer's output as (DTD eq. 10):
\begin{equation}
    R^\vh(\vx) =
    \frac{\partial R^y(\vh)}{\partial \vh} \Big|_{\tilde \vh = 0}
    \odot
    \vh  = \vh
\end{equation}
As $\vh$ is the ReLU output, we can write $R^\vh(\vx)$ also as (DTD eq. 11):
\begin{equation}
    \label{eq:dtd_linear_relevance_hidden}
    R^\vh(\vx) = [W \vx + \vb]^+
\end{equation}
The next step is to connect the input relevance $R^\vx(\vx)$ with the ReLU neurons' relevances $R^\vh$.
We will use \eqref{eq:dtd_recursive_taylor_step} and apply the Taylor
theorem to the relevance of each hidden neuron $\vh_\idx{j}$:
\begin{align}
    \label{eq:dtd_recursive_linear}
    R^\vx(\vx) &=
    \sum_{j=1}^d \left(
        \frac{\partial R_\idx{j}^\vh(\vx)}{\partial \vx}
        \Bigg|_{\vx = \tilde \vx^{(j)}}
        \odot
        (\vx - \tilde \vx^{(j)})
    \right)
    =
    \sum_{j=1}^d \left(
        \vw_j
        \odot
        (\vx - \tilde \vx^{(j)})
    \right),
\end{align}
where we used that the derivative of the hidden neuron $\vh_\idx{j}$ w.r.t the input is the weight vector $\vw_j = W_\idx{j:}$.

\paragraph{Relevance Propagation Rules}
For the root point $\tilde \vx$, we could, in theory, select any point in the
half-space $\vw_j \tilde \vx + b_j > 0$, as they are all valid according to the Taylor Theorem.
However, as it is beneficial to fully redistribute the relevance,
DTD
proposed selecting
a point that sets the $j$-th neuron relevance to zero, i.e., any point on the hyperplane $\vw_j^T \tilde \vx + \vb_j = 0$.
The non-differentiability at the ReLU hinge is resolved by picking the
gradient from the case $\vw_j \tilde \vx + b_j > 0$.

As there is no unique solution,
DTD derives different points by starting at the input $\vx$ and
moving along a direction $\vv_j$ such that $\tilde \vh_{j} = \vx - t \vv_j$ with
$t \in \R$.
The root point $\tilde \vx_j$ is then the intersection of the line $\vx + t
\vv_j$ with the hyperplane $\vw_j^T \vx + \vb_j = 0$.
Combining these equations
yields $t = - \frac{
\vw_j^T\vx + \vb_j
}{\vw_j^T\vv_j}
$ and therefore the root point $
\tilde \vx^{(j)} = \vx - \frac
    { \vw_j^T\vx + \vb_j}
    {\vw_j^T\vv_j}
    \vv_j
$.
Substituting this into \eqref{eq:dtd_recursive_linear} yields: (DTD-Appendix eq. 6, 7)
\begin{equation}
    \label{eq:dtd_linear_general_rule}
    R^\vx(\vx) =
    \sum_{j=1}^d \left(
        \vw_j \odot
            \frac{
                \vw_j^T\vx + \vb_j
            }{
                \vw_j^T\vv_j
            }
            \vv_j
    \right)
    =
    \sum_{j=1}^d \left(
    \vw_j \odot \frac{
            \vv_j
    }{
        \vw_j^T\vv_j
    }
        R_\idx{j}^\vh(\vx)
    \right).
\end{equation}

While almost all choices of $\vv$
yield a root point with $R^\vx(\tilde \vx) = 0$ (except $\vv \perp \vw$),
a few special directions exists:
\begin{itemize}
    \item \emph{The $w^2$-rule} chooses the closest root point in L2 metric: $\vv_j = \vw_j$. This will yield the root point $\tilde \vx = \vx - \frac{\vw_j}{\vw_j^T\vw_j} R^h(x)$ and the following relevance propagation rule: $R^x(\vx) = \sum_{j=1}^d
            \frac{
        \vw_j^2
            }{
                \vw_j^T\vw_j
            }
            R^\vh(\vx)
        $, where $\vw_j^2 = \vw_j \odot \vw_j$.
    \item \emph{The $z^+$ rule} uses a direction that always yields a positive root:
        $\vv_j = \ind_{\vw_j \ge 0} \vx$,
        which is preferred for positive inputs (e.g. from ReLU activations). The resulting
        root point is $\tilde \vx = \vx - \frac
            {\vw \ind_{\vw_j \ge 0} \vx }
            {\vw_j^T ( \ind_{\vw_j \ge 0} \vx)} R_\idx{j}^h(x)
        $ and the following relevance propagation rule: $
        R^x(\vx) = \sum_{j=1}^d 
            \frac{
                \vz_j^+
            }{
                \sum_i \vz_{ji}^+
            }
            R_\idx{j}^\vh(\vx)
        $, where $ \vz_j^+ = \ind_{\vw_j \ge 0} \vx \odot \vw_j^+  $.
    \item \emph{The gamma rule} proposed in \cite{montavon2019overview} uses the
    search direction $\vv_j\!=\!1 + \gamma \ind_{\vw_j \ge 0}\!\odot\!\vx$, where
    $\gamma \in \R^+$.  The corresponding relevance propagation rule is then:
    $ R^x(\vx) = \sum_{j=1}^d 
            \frac{
                \vw_j + \gamma \vz_j^+
            }{
                \vw_j^T (1 + \gamma \vz_j^+)
            }
            R_\idx{j}^\vh(\vx)
    $.
    In the limit $\gamma \to \infty$, the gamma rule becomes the $z^+$-rule.

    \item A special case is the \emph{
    \LRPO} rule which does not use any vector to find a root point
    but chooses $\tilde \vx = 0$. Although zero is not a
    valid root point in general, it was shown that
    \LRPO corresponds to gradient$\times$input
    \cite{avanti2016blackbox,ancona_towards_2018,kindermans_investigating_2016}.
    The \LRPe rule is an extension of \LRPO that adds a small $\varepsilon$ to
    increase numeric stability.  \end{itemize}
\newpage
\subsubsection{Which rule should be chosen?}
\label{subsec:which_rule_onelayer}

\begin{wrapfigure}{r}{0.52\textwidth}
\vspace{-1.0cm}
\centering
    \begin{subfigure}[t]{0.25\textwidth}
        \includegraphics{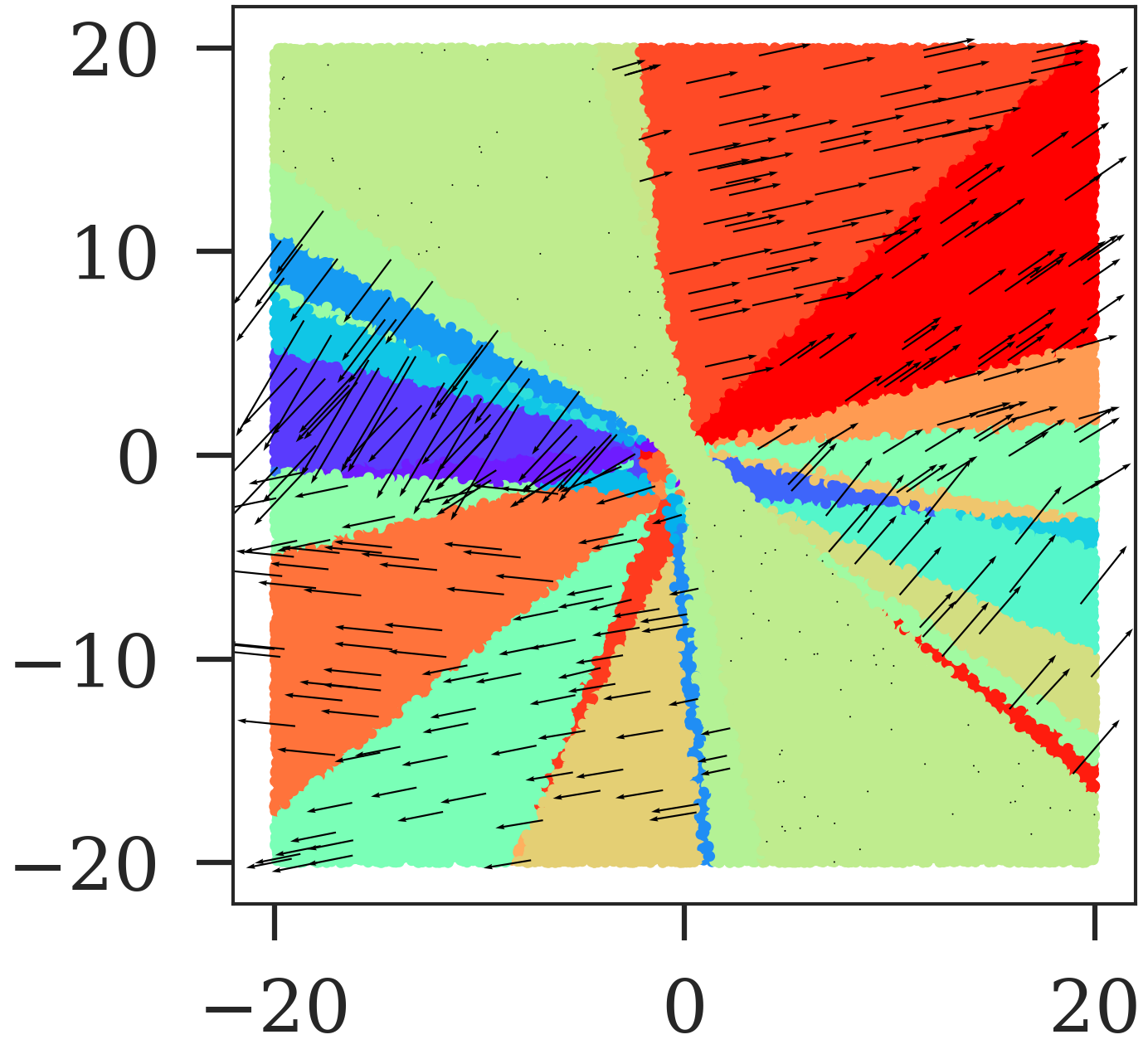}
        \caption{Non-positive biases}
    \end{subfigure}
    \begin{subfigure}[t]{0.25\textwidth}
        \includegraphics{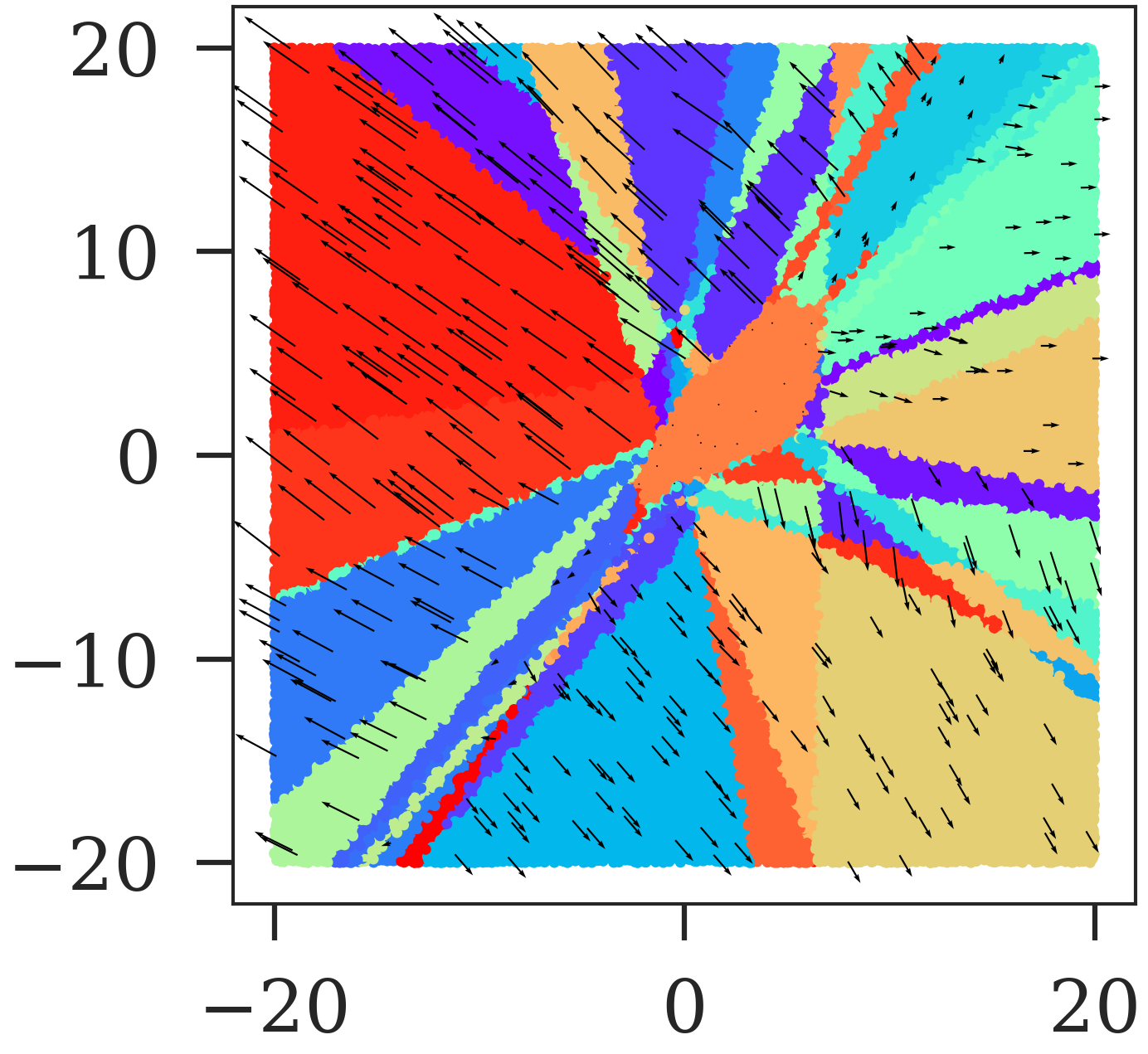}
        \caption{Unrestricted biases}
    \end{subfigure}
    \caption{Local linear regions of an randomly initialized neural network
    (3 layers, ReLU, 2 inputs, 10 hidden neurons).
    The biases are initialized
     \textbf{(a)}  non-positive and \textbf{(b)} unrestricted.
        The gradient are visualized as arrows for a random selection of points.
    }
    \label{fig:local_regions}
\end{wrapfigure}
The computed relevance values depend substantially on the rule. For example, the \LRPO rule
can compute negative relevance values, whereas the $z^+$ rule will always return
positive relevance values.
In \citet[Sec. 4.1, 4.2, 4.3]{montavon2017dtd}, the input domain
was used as the primary selection criterion.
For example, it was suggested to pick $z^+$ rule for $\R^+_0$ and the $w^2$ rule for the domain $\R$.
The input domain is not a sufficient selection for the root points, as it does not provide a unique solution.
In the later work \cite{montavon2019overview}, other selection criterion were proposed for deep neural networks, which we will analyze in Section \ref{subsec:which_root_points_deep}.
For now, we can conclude that no principled way to pick the roots and rules exists.

\subsubsection{Non-positive biases}
\label{subsub:dtd_restrict_bias}

In \cite{montavon2017dtd}, it was proposed to constrain the biases of the linear layer to be non-positive,
i.e. $b_j \le 0$. The main motivation was to
guarantee that the origin is a root point of the function $f$.
However, this is not the case, as the following simple counter-example will show.
Suppose the bias $b=-\vec{1}$. Then the function $f(\vec{0}) = 0$, as $[W\vec{0} - \vec{1}]^+ = 0$,
but the origin $\vec{0}$ is not a valid root point as the gradient is zero there. However,
any input $\vx$ with $f(\vx) \ge 0$ will have a non-zero gradient and will therefore be in a different local region.
In Figure \ref{fig:local_regions}, we visualized the local regions of a small 3-layered network
for non-positive and unrestricted bias.

\subsection{DTD for Deep Neural Neworks: The Training-Free Relevance Model}

Applying the recursive Taylor decomposition to a one-layered network
yielded a set of easily applied relevance propagation rules, which allowed to skip
computing the root points explicitly.
Of course, it would be desirable to skip the computation of roots
for deep neural networks too.
As solution, \cite{montavon2017dtd} proposed a so-called \emph{training-free} relevance
model. We follow the derivation from the review article \citep{montavon2018review}.

Let $R^{l+1}(\va_l)$ be the relevance computed for an upper layer.
\cite{montavon2017dtd} then makes the following assumption:
\begin{assumption}[Positive Linear Relevance]
The relevance of the upper layer $R^{l+1}(\va_{l+1})$
can be written as $R^{l+1}(\va_{l+1}) = \va_{l+1} \odot \vc_{l+1}$, where
$\vc_{l+1} \in \R^+$ should be a constant and positive vector.
\end{assumption}

As $R^{l+1}(\va_l) = \vc_{l+1} \odot [W_l \va_l + b_l]^+$, we can
construct a so-called relevance
neuron:
\begin{equation}
    \hat R^{l+1}(\va_{l+1}) = [\hat W_{l+1} \va_{l+1} + \hat \vb_{l+1}]^+,
\end{equation}
where we pulled $\vc$ into the layer's parameters:
$\hat W_{l+1} = W_{l+1} \odot C_{l+1}$ and where
$C_{l+1} = [\vc_{l+1}, \ldots, \vc_{l+1}] $ is a repeated version
of $\vc_{l+1}$ and $\hat \vb_{l+1} = \vb_{l+1}\odot \vc_{l+1}$.

This formulation is similar to the relevance of the hidden layer of the one-layer
network in \eqref{eq:dtd_linear_relevance_hidden}.
The difference is that the root point and search direction will
be based on the modified weights $\hat W_{l+1}$ and $\hat \vb_{l+1}$.
Using $\hat \vw_j = \hat W_{l+1_\idx{j:}} = \vc_{l+1_\idx{j}} W_{l+1_\idx{j:}}
$,
we can write the general relevance propagation rule
of \eqref{eq:dtd_linear_general_rule} as:
\begin{equation}
    \label{eq:dtd_linear_general_rule_mod}
    \hat R^{l}(\va_l) = \sum_{j=1}^d \left(
        \frac
            {\hat \vw_j\odot \vv_j}
            {\hat \vw_j^T\vv_j}
        \hat R^{l+1}_{\idx{j}}(f_l(\va_l))
    \right)
    =
    \sum_{j=1}^d \left(
        \frac
            {\vw_j \odot \vv_j}
            {\vw_j^T\vv_j}
        \hat R^{l+1}_{\idx{j}}(f_l(\va_l))
    \right),
\end{equation}
where the $\vc_{l+1_{j}}$ canceled out.
The corresponding root point would be: $\tilde \va_l^{(j)} =
    \va_l -
    \frac
        {\vw_j^T\vx + \vb_j}
        {\vw_j^T\vv_j}
    \vv_j
$.
Interestingly, this deviation recovers the one-layer case
from \eqref{eq:dtd_linear_general_rule}.
Thus, \cite{montavon2018review} argues that all the rules from the linear case (Section
\ref{subsec:dtd-single-layer}) can be applied to a deep neural network too.
This result can be easily extended to sum-pooling layers, as they are
equivalent to a linear layer with the weights of value 1.

\begin{remark}[Is it correct that $\vc_k$ is constant?]
A global constant $\vc_k$ cannot exist, as changing the input vector can result a in
totally different output, which would change the relevance magnitude.
A local approximation of $\vc_k$ could be correct
if root points stays within
the same local linear region where
the
function's gradient $\nabla f$ is locally constant.
\end{remark}

\begin{remark}[\textbf{C5}
no mechanism to enforce that the root point $\tilde \va_l^{(j)} \in N_{R^l}(\va_l)$?
]
The corresponding root point to
\eqref{eq:dtd_linear_general_rule_mod} would be: $
\tilde \va_l^{(j)} =
    \va_l -
    \frac
        {\vw_j^T\vx + \vb_j}
        {\vw_j^T\vv_j}
    \vv_j
$. Will this root point be in the local region $N_f(\va_l)$ of $\va_l$? Probably not, as
there is no mechanism enforcing this.
We test this in more detail in the empirical evaluation.
\end{remark}

\subsubsection{Which Root Points To Choose For A Deep Neural Network?}
\label{subsec:which_root_points_deep}

In Section \ref{subsec:which_rule_onelayer}, we already discussed that there is
no principled way to select the root points or corresponding rules.
For deep neural networks, DTD \cite{montavon2017dtd} originally proposed to pick the $z^+$-rule for all layers except
the first one.  In a more recent work \cite{montavon2019overview},
it was argued to use a combination of \LRPO, LRP-$\varepsilon$ and the
$\gamma$-rule. This was motivated by rather vague properties such
as the ``activations' entanglement'', ``spurious variations'', and ``spreading
of relevance''.
It is concluded that: ``\emph{Overall, in order to apply LRP successfully on a new task, it is important
to carefully inspect the properties of the neural network layers, and to ask the
human what kind of explanation is most understandable for him.}'' --
\citet[Sec. 10.3]{montavon2019overview}.
Thus, the choice of the rules lies in the hands of the user who might choose any
rule or root point.

\cite{kohlbrenner2020towards} introduced a similar combination of rules
as \emph{LRP-Composite}. For the convolution layers, they used the $z^+$-rule
(or the \LRPa) and for the fully-connected layers the \LRPO.
An improvement over using the
$z^+$-rule for all layers,
they found that this combination of rules did not suffer from
class-insensitivity, i.e., the saliency map do change when the explained output
class is changed.
However, it must be noted that this combination
relies on the particular properties of the convolutional neural network.
Specifically, there is little information mixing between more distant locations.
Furthermore, the explanations are still insensitive to the later convolutional layers:
the $z^+$ rule creates a fixed saliency map for the convolutional layers,
which, however, can be scaled by the output of the \LRPO-rule. For example,
if the final convolutional output has shape (8, 8) than the saliency map
can be scaled in an 8x8 grid.

\section{Analysis of the Recursive Application of the Taylor Theorem}
\label{sec:theoretic_analysis}

In the previous sections, we recapitulated how DTD applies the Taylor Theorem recursively
to a one-layer and a deep neural network,
and explained how the different propagation rules were derived.
In this section, we provide a theoretical analysis of the recursive application of the
Taylor Theorem. In particular, we study the
definition \ref{def:dtd_recursive_taylor} from Section \ref{subsec:dtd_recursive}.
As this definition is the most general formulation of the DTD theory,
we ensure that the results of our analysis
are applicable to all the propagation rules and
are also not caused by one specific approximation but are rather inherent to the
recursive application of the Taylor theorem.

The following propositions are proven in the Appendix \ref{appendix:proofs}.
The main idea of the proof is to apply the product rule to \eqref{eq:dtd_recursive_taylor_step}
and then analyze the individual terms.

\subsection{Size of admissable regions for the root points cannot be increased}
\begin{proposition}[\textbf{C1}: Recursively applying the Taylor Theorem cannot increase the size of admissible regions]
\label{prop:admissible_regions_root_points}
Given a ReLU network $f: \R^{d_1} \to \R_{\ge 0}^{d_{n+1}}$, recursive
relevance functions $R^l(\va_l)$ with $l \in \{1,\ldots,n\}$ according to definition
\ref{def:dtd_recursive_taylor}, and let $\xi$ index the explained logit,
then it holds for the admissible region $N_{R^l}(\va_l)$ for the root points $\tilde \va_l^{(j)}$
of the relevance function $R^l$ that
$N_{R^l}(\va_l) \subseteq N_{f_{n_\xi}\circ \ldots \circ f_{l}}(\va_l)$.
\end{proposition}

As the valid region for root points is restricted by the network $f$,
we
then we cannot evade the local region. This motivates a simple empirical
test in Section \ref{subsec:empirical_train_free_dtd}: for each root point, we can check whether it is contained in the correct
admissible region.
This result questions the motivation that the distance $|\vx-\tilde \vx|$ might be small
T from \ref{subsec:taylor_theorem_and_relevances}, as this distance remains bounded by the local
linear region of the network.

\subsection{Locally Constant Roots Imply Equivalence of Recursive Taylor and
Gradient×Input}
It is well known that \LRPO is equivalent to gradient$\times$input for ReLU networks.
This was first noted in \citep{avanti2016blackbox} and later also in
\citep{kindermans_investigating_2016,ancona_towards_2018}.
We proof the following generalization for the recursive application of
the Taylor Theorem in Appendix \ref{appendix:proofs_recursive_taylor}.

\begin{proposition}[\textbf{C2}]
\label{prop:recursive_taylor_equals_gradient}
Let $f: \R^{d_1} \to \R_{\ge 0}^{d_{n+1}}$ be a ReLU network,
$\xi$ be the index of the explained logit,
and $R^l(\va_l)$ (with $\,l \in 1\ldots n+1$) are recursive relevance functions according to definition \ref{def:dtd_recursive_taylor}.
If the root points $\tilde \va_l(\va_l)$ are locally constant w.r.t. the
layer's input ($
\,
\forall l \in {1\ldots n}:
\partial \tilde \va_l / \partial \va_l = 0
$),
then:
\begin{equation}
    R(\vx) = R(\tilde \vx) + \nabla f_\idx{\xi}(\vx) \odot (\vx - \tilde \vx),
\end{equation}
where $\vx = \va_1$ is the input vector and $R(\vx) = R^1(\vx)$.
\end{proposition}
The similarity with gradient$\times$input can be seen when choosing a root point
$\tilde \vx = \bm{0}$ such that $R(\bm{0})\!=\!\bm{0}$. Then, the resulting relevance would be $ \nabla f_\idx{\xi}(\vx) \odot
\vx$.

A fixed root point for each linear region
would be a valid and even desirable choice. For example, from an efficiency
perspective, it would be preferable to search for a valid root point in each
linear region only once.
Or one might want to select the one root point corresponding to the lowest
network output.
We also want to emphasize that no continuous constraint for selecting the
root points exists. Jumps at the boundaries between the linear region are allowed.
This result contradicts DTD's motivation described in Section
\ref{subsec:taylor_theorem_and_relevances}, as it explicitly aimed to find
something more ``stable'' than the gradient.

\subsection{Locally dependent root points}
As a next case, we will look at the more general case of root points depending
locally on the layer's input:
\begin{proposition}[\textbf{C3}]
\label{prop:recursive_taylor_relu}
For a ReLU network $f: \R^{d_1} \to \R_{\ge 0}^{d_{n+1}}$ with $n$ layers,
and layer activations $\va_l = f_{l-1}(\va_{l-1})$,
the relevance functions $R^{l-1}(a_{l-1}) $
of the recursive applications of the Taylor Theorem as given in \eqref{eq:dtd_recursive_taylor_step}
can be written as:
\begin{equation}
    \label{eq:dependent_root_points}
    R^{l-1}(a_{l-1}) =\!\sum_{j=1}^{d_{l}}
        \sum_{m = 1}^{d_{l+1}}
            \left[
            \left(
            \frac
                {\partial f_{l}(\va_{l}) }
                {\partial \va_{l_\idx{j}}}
                - \frac
                    {\partial \tilde \va_{l_\idx{j}}^{(m)}(\va_{l})}
                    {\partial \va_{l}}
                \cdot
                \frac
                    {\partial f_{l}(\va_{l}) }
                    {\partial \va_{l}}
            \right)
            \cdot
            \frac
                { \partial
                    R^{l+1}_\idx{m}\left(
                        f_{l}(\va_{l})
                    \right)
                }
                {
                    \partial f_{l}(\va_{l})
                }
            \right]
            \odot
            \left(
                \va_{l-1} - \tilde \va_{l-1}^{(j)}(\va_{l-1})
            \right)
    ,
\end{equation}
\end{proposition}

The relevance function $R^{l-1}$ is determined
by the next layer's Jacobian $\partial f_l(\va_l)/ \partial \va_l$,
and also a term including root point Jacobian $\partial \tilde \va_l^{(j)}/ \partial \va_l$.
Although some directions are recommended, the choice of root point is not
restricted per se.
It is merely recommended to choose it within the layer's input domain\footnote{%
In \citet[Section 4.1.]{montavon2017dtd},
the different rules were selected based on the input domain. However,
the $\gamma$ rule, introduced in a more recent work \cite{montavon2019overview},
can lead to root points outside the ReLU's input domain $\R^+$, e.g. for $\gamma = 0$
the root point is given by
$\tilde \vx = \vx - \frac{\vx}{\vw^T \vx} R^{l}(\vx)$
which can become negative for large relevance values $R^{l}(\vx)$.
}
and it should minimize the explained relevance.
Any root point could be chosen, as long as it is from the linear region $N_{R^l_\idx{k}}(\va_l)$.
However, this also means that $R^{l-1}(\va_{l-1})$
can be influenced arbitrarily by the root point's Jacobian.
Therefore, any explanation could be justified.
A theory under which anything can be justified is clearly insufficient.

\subsection{Why Not Use Analytic Activation Functions (Softplus)?}

For ReLu networks, the Deep Taylor Decomposition suffers from the problem that
the root point must be from the local linear region around the layer input $\va_l$.
A possible solution would be to use an analytic activation function,
e.g. the Softplus activation.
This would allow to choose any root point in
$\R^{d_{l}}$, although a sufficiently good approximation might require
an unreasonable amount of higher-order terms.
The main obstacle would be that with each decomposition,
higher-order derivatives are accumulated:
\begin{proposition}[\textbf{C4}]
\label{prop:higher_order_terms}
Let
$f: \R^{d_1} \mapsto \R^{d_{n+1}}$
be
a neural network,
contains an analytic activation function,
then each recursive application of the Taylor Theorem
yields a higher-order derivative of the form:
\begin{equation}
        \frac
        {\partial}
        {\partial \va_l}
        \cdot
        \left[
        \frac
            {
                \partial R^{l+1}_\idx{j}\left(
                    f_{l}(\va_{l})
                \right)
            }
            {\partial \va_{l}}
        \right]_{\va_{l} =\tilde \va_l^{(j)}(\va_{l})}
        \cdot
        \left\{\va_l - \tilde \va_{l}^{(j)}(\va_{l})\right\}_\idx{k}.
\end{equation}
\end{proposition}
Thus, for a n-layered network, we would get n-ordered derivatives.
The problem is that it is unclear how these chains of higher-order derivatives
behave.

\section{Experiments}
\label{sec:experiments}

\subsection{ (C6) DTD-Train-Free}
\label{subsec:empirical_train_free_dtd}

\begin{table}[t]
    \centering
    \caption{
        Empirical results of different DTD rules on a small neural network (3 layers, 10 input dimensions, 10 hidden dimensions).
        They show that the root points picked by the rules are not within the local region
        of the input, as each rule produced outputs below 100\%.
        It is also the case that some root points will have the exact same network output as the original
        input.
    }
    \label{tab:dtd_rules_empricial_results}

    \begin{tabular}{lllll}
    \toprule
    Evaluation \textbackslash~Rule & LRP$_{0}$ & $\gamma=1.0$ &   $w^2$ &   $z^+$ \\
    \midrule
    Same local linear region [expected 100\%]    &    41.20\% &       38.70\% &  37.70\% &  41.10\% \\
    Same network output [expected 0\%] &    14.62\% &       13.85\% &  14.01\% &  13.99\% \\
    \bottomrule
    \end{tabular}
\end{table}

We implemented the train-free DTD using an explicit computation of the root points.
The network consists of 3 linear layers, each with a ReLU activation. The input and each layer has 10 dimensions.
We initialized the network with random weights and
used non-positive biases, as
\cite{montavon2017dtd} suggested (even though we have shown that this has not the same
consequences as claimed in \cite{montavon2017dtd}.
As we are
only interested in disproving claims,
it is acceptable to show that there exists one
of neural network on which the DTD delivers inconsistent results.
Therefore, we also did not train the network on any task.

We provide pseudocode for our implementation in Algorithm \ref{alg:dtd_train_free}.
The main simplifications of the implementation are that (1) the relevance of the higher layers is computed with the input of the layer and not the root point,
and (2) the root points are computed using the search directions outlined in section \ref{subsec:dtd-single-layer}.
We tested our implementation against Captum's implementation of the DTD
\citep{kokhlikyan2020captum} and found the deviation to be less than \num{1e-8}.

Verifying that the two points are within the local region would
require to show that the gradients are equal and that there is a path between the two points
with all points on the path also having equal gradients.
As the last part is more difficult to show, we only test the
necessary condition of equal gradients.
Therefore, we compare the gradient of the input with the gradient of the root points
$ |\nabla f(\vx) - \nabla f(\tilde \vx)|$
on 1000 random inputs. The input points were sampled such that it has a network
output greater than \num{0.1}.

We reported the numerical results in Table \ref{tab:dtd_rules_empricial_results}.
Less than 100\% of all root points have gradient differences that are zero, thus
root point exists which must be from a different local region.
This violates Proposition \ref{prop:admissible_regions_root_points}, which requires
all root points to be within the function's local region.
Although we only show results on an exemplary 3-layered network, the situation would only be worse
for more complex networks as the number of local regions increases exponentially with layer depth \citep{montufar_number_2014}.

As a second analysis, we tested how the root points influence the network output.
One might assume that a root point will alter the network output.
However, this is not always the case (see row ``Same network output'' in Table
\ref{tab:dtd_rules_empricial_results}).
At least, these root points do not
explain the output of the neural network.

\subsection{(C7) Applying Sanity Checks to (Arras et al., 2022)}

\begin{figure}[t]
    \centering
    \vspace{-1.2cm}
    \import{./figures/saliency/}{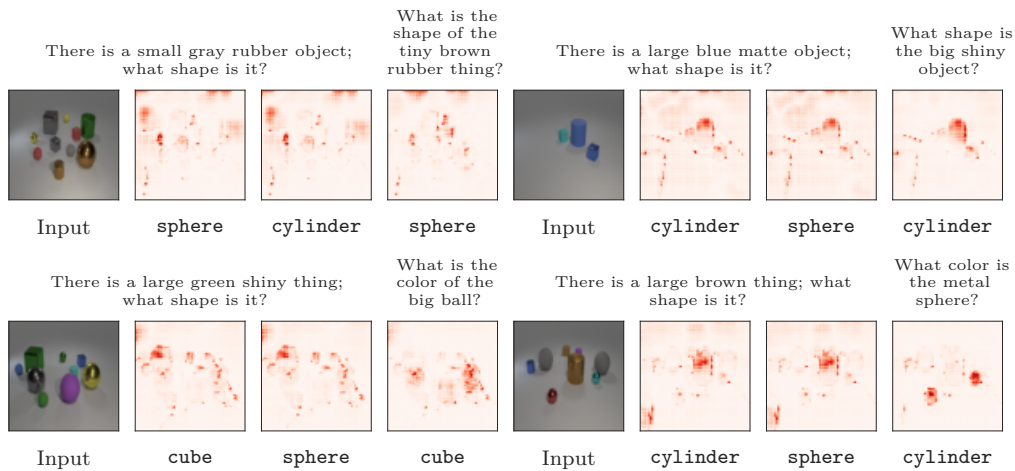}
    \caption{
        (1) Input images from the CLEVER-XAI dataset with the \LRPa saliency maps
        computed for the (2) correct class, (3) an incorrect class, and (4) a different question.
        The original question is written above. The saliency maps do not change visually when
        a different output class is explained. However, changing the question
        highlights other regions.
    }
    \label{fig:sanity_checks}
\end{figure}

\begin{wrapfigure}{r}{0.38\textwidth}
    \centering
    \vspace{-0.0cm}
    \import{./figures/}{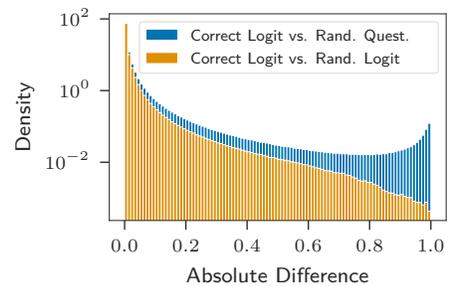}
    \caption{
        The histogram of absolute differences between the saliency maps for \emph{Correct Logit vs. Random Question}
        and \emph{Correct Logit vs. Random Logit}.
    }
    \label{fig:hist_abs_diffs_saliency_maps}
\end{wrapfigure}
A recent work \citep{arras2022clevr} evaluated different saliency methods
on the CLEVR VQA dataset using ground truth segmentation masks.
Interestingly, they found \LRPa (equivalent to the DTD $z^+$-rule)
to highlight the object of interest particularly well:
\emph{``[...] a high connection between the relevance heatmaps
and the target objects of each question''}.
This finding seems to
contradict \cite{sixt2020wel}, which found that \LRPa becomes
independent of the network's deeper layer.
In \cite{arras2022clevr}, it was therefore concluded: \emph{``Maybe the phenomenon described in
\citep{sixt2020wel} becomes predominant in the asymptotic case of a neural
network with a very high number of layers, [...]''}.

A simple empirical test would have been to check if \LRPa's saliency maps
change when the network's last layer is changed.
To perform this test, we replicated their setup and trained a relation network
\citep{santoro2017rn} on the CLEVR V1.0 dataset \cite{johnson2017clevr}.
The network reached an accuracy of
93.47\%, comparable to 93.3\% \citep{arras2022clevr} and 95.5\% \citep{santoro2017rn}.
We included more details about the model in Appendix \ref{app:clevr_model}.

We then compared \num{1000} \LRPa's saliency maps for the correct answer, an
incorrect answer (but from the same category), and the correct answer but a
different question.
It is valid to ask for an explanation of a different class, for example, to
understand which evidence is present for \texttt{sphere} instead of \texttt{cube}.
The saliency maps were scaled to cover the range [0,1], and the differences were measured using the mean absolute difference.
In Figure \ref{fig:hist_abs_diffs_saliency_maps}, a histogram of the differences is shown.
While the saliency maps are very similar in both cases,
there seems to be more variability in the question: for ``correct logit vs. random question'' there is an order of magnitude more pixels with a difference of $\approx 1$.
When looking at the resulting saliency maps in Figure \ref{fig:sanity_checks},
one can see that the saliency maps differ quite significantly when changing the question.
In contrast, the saliency map of the wrong answer does not change.

First, these results validate the claim in \citep{sixt2020wel} that
\LRPa is independent of the network's deeper layer. Second, they
indicate that an information leakage between the question and
\LRPa's saliency maps is present.

The reason for this information leakage can be found in the specific
architecture of the relation networks: pairs are formed between all feature-map locations
of the convolutional output. As the feature map has shape $(8, 8, 24)$,
$64\cdot 64$ pairs are formed (i.e., $\text{height}^2 \cdot \text{width}^2$).
Additionally, the question embedding
produced by an LSTM layer is concatenated to each pair.
This yields triples of $(\vv_{ij},
\vv_{kl}, \vo_\text{LSTM})$, where $i,j,k,l \in \{1,\ldots, 8\}$, and $\vv \in
\R^{8\times8\times24}$ is the convolutional stack's output.
As the convolutional layers and the LSTM layer are trained together,
their representations are aligned. Thus, changing the LSTM embedding will
change the internal representation in the subsequent layers.
For relevance locations, the question embedding will
match better with the convolutional activation and, therefore will
lead to a higher saliency map at relevant locations.
However, the saliency maps will still become independent of the network's deeper layer.

The implications are substantial: for example,
if the model's final layers' were fine-tuned on a
new task, the \LRPa explanation would not change and could not be used to explain
this model. Even worse, if your model was altered to predict spheres
instead of cubes, the LRP explanation would not reflect this.

It is quite fascinating that the \LRPa explanations highlight the right object
according to the ground truth, but fail to highlight evidence for the wrong object.
This result also shows how difficult it is to evaluate explanation methods
empirically.

\section{Conclusion}
\label{sec:conclusion}

We have shown that DTD, which has been cited as the theoretical foundation of numerous follow-up post-hoc explanation techniques
\citep{ali2022transformers,binder_layer-wise_2016, kindermans_learning_2018,arras2017explaining,hui_batchnorm_2019,huber2019enhancing,eberle2020building},
exhibits serious flaws that explain why saliency maps created with these methods are independent of the output.
From \cite{sixt2020wel},
we know that
that the positive matrices produced by the $z^+$-rule will converge to a rank-1
matrix. These positive matrices stem from a specific selection of the root-point, and as the selection of the root-points is not restricted,
the $z^+$-rule can be justified by the DTD theory, as every other
explanation could be - by picking an appropriate root.

DTD as a theoretical framework for explanations is under-constraint, and can be considered insufficient. Caution must be used when using explanations derived from this theory.
At the core of the problem, there is no restriction and little guidance on choosing the root points.
Under certain conditions (constant root points), DTD reduces to backpropagating the gradient, albeit hidden behind a complex mathematical structure.
In the other case (input-dependent root points), DTD leaves open a backdoor through which virtually any explanation can be created by crafting the root point's Jacobian accordingly.
However, this again is obfuscated by the theory rather than made transparent.

Since its first ArXiv submission \citep{montavon2015explaining}, the DTD publication has been cited numerous times. Although even the authors have reported class-insensitive behavior \citep{kohlbrenner2020towards,montavon2018review}\footnote{
\emph{``A reduced number of fully-connected layers avoids that the relevance, when redistributed backwards, looses its connection to the concept being predicted.''} -- \citet[Sec 5.1.]{montavon2018review}
},
follow-up works have readily used DTD's key concepts, motivated by the seemingly robust mathematical foundation, instead of searching for the underlying reasons.
Furthermore, explanations based on DTD were used in various
applications, for example, for validating their model
\citep{andresen2020towards}, gain insights into geoscientific questions \citep{toms2020physically}, or conduct user studies
\citep{alqaraawi2020eval}.

While we were able to discover serious issues of DTD, we
do not see a solution how to solve them.
We therefore want to point out that other theoretically well-justified methods exist:
\emph{(Deletion of Information)} which information can be deleted without changing
the network output? One approach uses noise \citep{schulz2020Restricting}, other discrete deletion
of the input \citep{macdonald2019ratedistortion}.
\emph{(Testing prediction capabilities)}:
We can test whether certain concepts are present in the network by
trying to predict them from intermediate features \citep{kim2018interpretability}.
\emph{(Model inversion)}: How would the input need to change to predict another class?
This question can be answered using invertible models
\citep{Hvilshj2021ECINNEC,dombrowski2021diffeomorphic} or conditional
generative models \citep{Singla2020Explanation}.
\emph{(Simple Models)}: If a similar performance is achieved by a simpler, more
interpretable model, why not simply use that? For example,
\citep{zhang2021invertible} replaced part of the network with a linear model.
All these approaches do not come with a complicated mathematical
superstructure, rather they have a simple and intuitive motivation.

\subsubsection*{Broader Impact Statement}

Although our work focuses on the theoretical foundations of a particular
explanation method, we see broader implications of this work.
Our work demonstrates that the theoretical foundation of explanation methods need rigorous analysis before they can support the trust that 
developers, users, and even regulatory bodies may put in it.
This is especially important in the field of explainable AI since empirically evaluating explanations is difficult.  

The field offers a variety of explanation methods, and ways to test the quality of explanations. We recommend using more than just one method and employing a range of metrics and user tests to make sure explanations are helpful in potentially critical use-cases such as medical decision making or the screening of job applications. 

\subsubsection*{Acknowledgement}

We want to thank the reviewers for their helpful feedback that improved this manuscript further. The computation were done on the Curta cluster provided by the Zedat, Freie Universtität Berlin \citep{Bennett2020}. 
Finally, we thank Jonas Köhler for discussions about the manuscript idea.

\newpage

\newpage

\bibliography{tmlr}

\begin{thebibliography}{56}
\providecommand{\natexlab}[1]{#1}
\providecommand{\url}[1]{\texttt{#1}}
\expandafter\ifx\csname urlstyle\endcsname\relax
  \providecommand{\doi}[1]{doi: #1}\else
  \providecommand{\doi}{doi: \begingroup \urlstyle{rm}\Url}\fi

\bibitem[Ali et~al.(2022)Ali, Schnake, Eberle, Montavon, M{\"u}ller, and
  Wolf]{ali2022transformers}
Ameen Ali, Thomas Schnake, Oliver Eberle, Gr{\'e}goire Montavon, Klaus-Robert
  M{\"u}ller, and Lior Wolf.
\newblock {XAI} for transformers: Better explanations through conservative
  propagation.
\newblock In Kamalika Chaudhuri, Stefanie Jegelka, Le~Song, Csaba Szepesvari,
  Gang Niu, and Sivan Sabato (eds.), \emph{Proceedings of the 39th
  International Conference on Machine Learning}, volume 162 of
  \emph{Proceedings of Machine Learning Research}, pp.\  435--451. PMLR, 17--23
  Jul 2022.
\newblock URL \url{https://proceedings.mlr.press/v162/ali22a.html}.

\bibitem[Alqaraawi et~al.(2020)Alqaraawi, Schuessler, Wei\ss{}, Costanza, and
  Berthouze]{alqaraawi2020eval}
Ahmed Alqaraawi, Martin Schuessler, Philipp Wei\ss{}, Enrico Costanza, and
  Nadia Berthouze.
\newblock Evaluating saliency map explanations for convolutional neural
  networks: A user study.
\newblock In \emph{Proceedings of the 25th International Conference on
  Intelligent User Interfaces}, IUI '20, pp.\  275–285, New York, NY, USA,
  2020. Association for Computing Machinery.
\newblock ISBN 9781450371186.
\newblock \doi{10.1145/3377325.3377519}.
\newblock URL \url{https://doi.org/10.1145/3377325.3377519}.

\bibitem[Ancona et~al.(2018)Ancona, Ceolini, Öztireli, and
  Gross]{ancona_towards_2018}
Marco Ancona, Enea Ceolini, Cengiz Öztireli, and Markus Gross.
\newblock Towards better understanding of gradient-based attribution methods
  for {Deep} {Neural} {Networks}.
\newblock In \emph{International Conference on Learning Representations},
  February 2018.
\newblock URL \url{https://openreview.net/forum?id=Sy21R9JAW}.

\bibitem[Andresen et~al.(2020)Andresen, W{\"o}llhaf, Hohlbaum, Lewejohann,
  Hellwich, Th{\"o}ne-Reineke, and Belik]{andresen2020towards}
Niek Andresen, Manuel W{\"o}llhaf, Katharina Hohlbaum, Lars Lewejohann, Olaf
  Hellwich, Christa Th{\"o}ne-Reineke, and Vitaly Belik.
\newblock Towards a fully automated surveillance of well-being status in
  laboratory mice using deep learning: Starting with facial expression
  analysis.
\newblock \emph{PLoS One}, 15\penalty0 (4):\penalty0 e0228059, 2020.

\bibitem[Arora et~al.(2018)Arora, Basu, Mianjy, and
  Mukherjee]{arora_understanding_2018}
Raman Arora, Amitabh Basu, Poorya Mianjy, and Anirbit Mukherjee.
\newblock Understanding {Deep} {Neural} {Networks} with {Rectified} {Linear}
  {Units}.
\newblock In \emph{International Conference on Learning Representations
  (ICLR)}, February 2018.
\newblock URL \url{https://openreview.net/forum?id=B1J_rgWRW}.

\bibitem[Arras et~al.(2017)Arras, Montavon, M{\"u}ller, and
  Samek]{arras2017explaining}
Leila Arras, Gr{\'e}goire Montavon, Klaus-Robert M{\"u}ller, and Wojciech
  Samek.
\newblock Explaining recurrent neural network predictions in sentiment
  analysis.
\newblock \emph{EMNLP 2017}, pp.\  159, 2017.

\bibitem[Arras et~al.(2022)Arras, Osman, and Samek]{arras2022clevr}
Leila Arras, Ahmed Osman, and Wojciech Samek.
\newblock Clevr-xai: A benchmark dataset for the ground truth evaluation of
  neural network explanations.
\newblock \emph{Information Fusion}, 81:\penalty0 14--40, 2022.
\newblock ISSN 1566-2535.
\newblock \doi{https://doi.org/10.1016/j.inffus.2021.11.008}.
\newblock URL
  \url{https://www.sciencedirect.com/science/article/pii/S1566253521002335}.

\bibitem[Bach et~al.(2015)Bach, Binder, Montavon, Klauschen, Müller, and
  Samek]{bach_pixel-wise_2015}
Sebastian Bach, Alexander Binder, Grégoire Montavon, Frederick Klauschen,
  Klaus~Robert Müller, and Wojciech Samek.
\newblock On pixel-wise explanations for non-linear classifier decisions by
  layer-wise relevance propagation.
\newblock \emph{PLoS ONE}, 10\penalty0 (7), 2015.
\newblock ISSN 19326203.
\newblock \doi{10.1371/journal.pone.0130140}.
\newblock URL
  \url{http://journals.plos.org/plosone/article/file?id=10.1371/journal.pone.0130140&type=printable}.
\newblock 00067.

\bibitem[Balduzzi et~al.(2017)Balduzzi, Frean, Leary, Lewis, Ma, and
  McWilliams]{balduzzi_shattered_2017}
David Balduzzi, Marcus Frean, Lennox Leary, J.~P. Lewis, Kurt Wan-Duo Ma, and
  Brian McWilliams.
\newblock The {Shattered} {Gradients} {Problem}: {If} resnets are the answer,
  then what is the question?
\newblock In \emph{Proceedings of the 34th {International} {Conference} on
  {Machine} {Learning}}, pp.\  342--350. PMLR, July 2017.
\newblock URL \url{https://proceedings.mlr.press/v70/balduzzi17b.html}.
\newblock ISSN: 2640-3498.

\bibitem[Bennett et~al.(2020)Bennett, Melchers, and Proppe]{Bennett2020}
Loris Bennett, Bernd Melchers, and Boris Proppe.
\newblock Curta: A general-purpose high-performance computer at zedat, freie
  universität berlin.
\newblock http://dx.doi.org/10.17169/refubium-26754, 2020.

\bibitem[Binder et~al.(2016)Binder, Montavon, Lapuschkin, Müller, and
  Samek]{binder_layer-wise_2016}
Alexander Binder, Grégoire Montavon, Sebastian Lapuschkin, Klaus-Robert
  Müller, and Wojciech Samek.
\newblock Layer-{Wise} {Relevance} {Propagation} for {Neural} {Networks} with
  {Local} {Renormalization} {Layers}.
\newblock In \emph{Artificial {Neural} {Networks} and {Machine} {Learning} –
  {ICANN} 2016}, Lecture {Notes} in {Computer} {Science}, pp.\  63--71.
  Springer International Publishing, 2016.
\newblock ISBN 978-3-319-44781-0.
\newblock \doi{10.1007/978-3-319-44781-0_8}.

\bibitem[Chen et~al.(2019)Chen, Song, Wainwright, and Jordan]{chen2018lshapley}
Jianbo Chen, Le~Song, Martin~J. Wainwright, and Michael~I. Jordan.
\newblock L-shapley and c-shapley: Efficient model interpretation for
  structured data.
\newblock In \emph{International Conference on Learning Representations}, 2019.
\newblock URL \url{https://openreview.net/forum?id=S1E3Ko09F7}.

\bibitem[Dombrowski et~al.(2019)Dombrowski, Alber, Anders, Ackermann,
  M{\"u}ller, and Kessel]{dombrowski2019explanations}
Ann-Kathrin Dombrowski, Maximillian Alber, Christopher Anders, Marcel
  Ackermann, Klaus-Robert M{\"u}ller, and Pan Kessel.
\newblock Explanations can be manipulated and geometry is to blame.
\newblock \emph{Advances in Neural Information Processing Systems}, 32, 2019.

\bibitem[Dombrowski et~al.(2021)Dombrowski, Gerken, and
  Kessel]{dombrowski2021diffeomorphic}
Ann-Kathrin Dombrowski, Jan~E Gerken, and Pan Kessel.
\newblock Diffeomorphic explanations with normalizing flows.
\newblock In \emph{ICML Workshop on Invertible Neural Networks, Normalizing
  Flows, and Explicit Likelihood Models}, 2021.
\newblock URL \url{https://openreview.net/forum?id=ZBR9EpEl6G4}.

\bibitem[Eberle et~al.(2020)Eberle, B{\"u}ttner, Kr{\"a}utli, M{\"u}ller,
  Valleriani, and Montavon]{eberle2020building}
Oliver Eberle, Jochen B{\"u}ttner, Florian Kr{\"a}utli, Klaus-Robert
  M{\"u}ller, Matteo Valleriani, and Gr{\'e}goire Montavon.
\newblock Building and interpreting deep similarity models.
\newblock \emph{IEEE Transactions on Pattern Analysis and Machine
  Intelligence}, 2020.

\bibitem[Folland(2002)]{folland_advanced_2002}
G.~B. Folland.
\newblock \emph{Advanced calculus}.
\newblock Prentice Hall, Upper Saddle River, NJ, 2002.
\newblock ISBN 978-0-13-065265-2.

\bibitem[Holzinger et~al.(2022)Holzinger, Saranti, Molnar, Biecek, and
  Samek]{holzinger2022explainable}
Andreas Holzinger, Anna Saranti, Christoph Molnar, Przemyslaw Biecek, and
  Wojciech Samek.
\newblock \emph{Explainable AI Methods - A Brief Overview}, pp.\  13--38.
\newblock Springer International Publishing, Cham, 2022.
\newblock ISBN 978-3-031-04083-2.
\newblock \doi{10.1007/978-3-031-04083-2_2}.
\newblock URL \url{https://doi.org/10.1007/978-3-031-04083-2_2}.

\bibitem[Huber et~al.(2019)Huber, Schiller, and Andr{\'e}]{huber2019enhancing}
Tobias Huber, Dominik Schiller, and Elisabeth Andr{\'e}.
\newblock Enhancing explainability of deep reinforcement learning through
  selective layer-wise relevance propagation.
\newblock In \emph{Joint German/Austrian Conference on Artificial Intelligence
  (K{\"u}nstliche Intelligenz)}, pp.\  188--202. Springer, 2019.

\bibitem[Hui \& Binder(2019)Hui and Binder]{hui_batchnorm_2019}
Lucas Y.~W. Hui and Alexander Binder.
\newblock {BatchNorm} {Decomposition} for {Deep} {Neural} {Network}
  {Interpretation}.
\newblock In \emph{Advances in {Computational} {Intelligence}}, volume 11507,
  pp.\  280--291. Springer International Publishing, 2019.
\newblock \doi{10.1007/978-3-030-20518-8_24}.
\newblock URL \url{http://link.springer.com/10.1007/978-3-030-20518-8_24}.
\newblock Series Title: Lecture Notes in Computer Science.

\bibitem[Hvilsh{\o}j et~al.(2021)Hvilsh{\o}j, Iosifidis, and
  Assent]{Hvilshj2021ECINNEC}
Frederik Hvilsh{\o}j, Alexandros Iosifidis, and Ira Assent.
\newblock Ecinn: Efficient counterfactuals from invertible neural networks.
\newblock \emph{ArXiv}, abs/2103.13701, 2021.

\bibitem[Johnson et~al.(2017)Johnson, Hariharan, Van Der~Maaten, Fei-Fei,
  Lawrence~Zitnick, and Girshick]{johnson2017clevr}
Justin Johnson, Bharath Hariharan, Laurens Van Der~Maaten, Li~Fei-Fei,
  C~Lawrence~Zitnick, and Ross Girshick.
\newblock Clevr: A diagnostic dataset for compositional language and elementary
  visual reasoning.
\newblock In \emph{Proceedings of the IEEE conference on computer vision and
  pattern recognition}, pp.\  2901--2910, 2017.

\bibitem[Kim et~al.(2018)Kim, Wattenberg, Gilmer, Cai, Wexler, Viegas,
  et~al.]{kim2018interpretability}
Been Kim, Martin Wattenberg, Justin Gilmer, Carrie Cai, James Wexler, Fernanda
  Viegas, et~al.
\newblock Interpretability beyond feature attribution: Quantitative testing
  with concept activation vectors (tcav).
\newblock In \emph{International conference on machine learning}, pp.\
  2668--2677. PMLR, 2018.

\bibitem[Kindermans et~al.(2016)Kindermans, Schütt, Müller, and
  Dähne]{kindermans_investigating_2016}
Pieter-Jan Kindermans, Kristof Schütt, Klaus-Robert Müller, and Sven Dähne.
\newblock Investigating the influence of noise and distractors on the
  interpretation of neural networks.
\newblock Technical Report arXiv:1611.07270, arXiv, November 2016.
\newblock URL \url{http://arxiv.org/abs/1611.07270}.
\newblock arXiv:1611.07270 [cs, stat] type: article.

\bibitem[Kindermans et~al.(2018)Kindermans, Schütt, Alber, Müller, Erhan,
  Kim, and Dähne]{kindermans_learning_2018}
Pieter-Jan Kindermans, Kristof~T. Schütt, Maximilian Alber, Klaus-Robert
  Müller, Dumitru Erhan, Been Kim, and Sven Dähne.
\newblock Learning how to explain neural networks: {PatternNet} and
  {PatternAttribution}.
\newblock February 2018.
\newblock URL \url{https://openreview.net/forum?id=Hkn7CBaTW}.

\bibitem[Kohlbrenner et~al.(2020)Kohlbrenner, Bauer, Nakajima, Binder, Samek,
  and Lapuschkin]{kohlbrenner2020towards}
Maximilian Kohlbrenner, Alexander Bauer, Shinichi Nakajima, Alexander Binder,
  Wojciech Samek, and Sebastian Lapuschkin.
\newblock Towards best practice in explaining neural network decisions with
  lrp.
\newblock In \emph{2020 International Joint Conference on Neural Networks
  (IJCNN)}, pp.\  1--7. IEEE, 2020.

\bibitem[Kokhlikyan et~al.(2020)Kokhlikyan, Miglani, Martin, Wang, Alsallakh,
  Reynolds, Melnikov, Kliushkina, Araya, Yan, and
  Reblitz-Richardson]{kokhlikyan2020captum}
Narine Kokhlikyan, Vivek Miglani, Miguel Martin, Edward Wang, Bilal Alsallakh,
  Jonathan Reynolds, Alexander Melnikov, Natalia Kliushkina, Carlos Araya, Siqi
  Yan, and Orion Reblitz-Richardson.
\newblock Captum: A unified and generic model interpretability library for
  pytorch, 2020.

\bibitem[Kumar et~al.(2020)Kumar, Venkatasubramanian, Scheidegger, and
  Friedler]{kumar2020problems}
I.~Elizabeth Kumar, Suresh Venkatasubramanian, Carlos Scheidegger, and Sorelle
  Friedler.
\newblock Problems with shapley-value-based explanations as feature importance
  measures.
\newblock In \emph{Proceedings of the 37th International Conference on Machine
  Learning}, volume 119 of \emph{Proceedings of Machine Learning Research},
  pp.\  5491--5500. PMLR, 2020.
\newblock URL \url{https://proceedings.mlr.press/v119/kumar20e.html}.

\bibitem[Lundberg \& Lee(2017)Lundberg and Lee]{lundberg2017unified}
Scott~M Lundberg and Su-In Lee.
\newblock A unified approach to interpreting model predictions.
\newblock \emph{Advances in neural information processing systems}, 30, 2017.

\bibitem[Lundstrom et~al.(2022)Lundstrom, Huang, and
  Razaviyayn]{lundstrom2022int_grad}
Daniel~D Lundstrom, Tianjian Huang, and Meisam Razaviyayn.
\newblock A rigorous study of integrated gradients method and extensions to
  internal neuron attributions.
\newblock In Kamalika Chaudhuri, Stefanie Jegelka, Le~Song, Csaba Szepesvari,
  Gang Niu, and Sivan Sabato (eds.), \emph{Proceedings of the 39th
  International Conference on Machine Learning}, volume 162 of
  \emph{Proceedings of Machine Learning Research}, pp.\  14485--14508. PMLR,
  17--23 Jul 2022.
\newblock URL \url{https://proceedings.mlr.press/v162/lundstrom22a.html}.

\bibitem[Macdonald et~al.(2019)Macdonald, Wäldchen, Hauch, and
  Kutyniok]{macdonald2019ratedistortion}
Jan Macdonald, Stephan Wäldchen, Sascha Hauch, and Gitta Kutyniok.
\newblock A rate-distortion framework for explaining neural network decisions,
  2019.

\bibitem[Montavon et~al.(2015)Montavon, Bach, Binder, Samek, and
  M{\"u}ller]{montavon2015explaining}
Gr{\'e}goire Montavon, Sebastian Bach, Alexander Binder, Wojciech Samek, and
  Klaus-Robert M{\"u}ller.
\newblock Explaining nonlinear classification decisions with deep taylor
  decomposition.
\newblock \emph{arXiv preprint arXiv:1512.02479}, 2015.

\bibitem[Montavon et~al.(2019)Montavon, Binder, Lapuschkin, Samek, and
  M{\"u}ller]{montavon2019overview}
Gr{\'e}goire Montavon, Alexander Binder, Sebastian Lapuschkin, Wojciech Samek,
  and Klaus-Robert M{\"u}ller.
\newblock Layer-wise relevance propagation: an overview.
\newblock \emph{Explainable AI: interpreting, explaining and visualizing deep
  learning}, pp.\  193--209, 2019.

\bibitem[Montavon et~al.(2017)Montavon, Lapuschkin, Binder, Samek, and
  Müller]{montavon2017dtd}
Grégoire Montavon, Sebastian Lapuschkin, Alexander Binder, Wojciech Samek, and
  Klaus-Robert Müller.
\newblock Explaining nonlinear classification decisions with deep taylor
  decomposition.
\newblock \emph{Pattern Recognition}, 65:\penalty0 211--222, 2017.
\newblock ISSN 0031-3203.
\newblock \doi{https://doi.org/10.1016/j.patcog.2016.11.008}.
\newblock URL
  \url{https://www.sciencedirect.com/science/article/pii/S0031320316303582}.

\bibitem[Montavon et~al.(2018)Montavon, Samek, and Müller]{montavon2018review}
Grégoire Montavon, Wojciech Samek, and Klaus-Robert Müller.
\newblock Methods for interpreting and understanding deep neural networks.
\newblock \emph{Digital Signal Processing}, 73:\penalty0 1--15, February 2018.
\newblock ISSN 10512004.
\newblock \doi{10.1016/j.dsp.2017.10.011}.
\newblock URL
  \url{https://linkinghub.elsevier.com/retrieve/pii/S1051200417302385}.

\bibitem[Montufar et~al.(2014)Montufar, Pascanu, Cho, and
  Bengio]{montufar_number_2014}
Guido~F Montufar, Razvan Pascanu, Kyunghyun Cho, and Yoshua Bengio.
\newblock On the {Number} of {Linear} {Regions} of {Deep} {Neural} {Networks}.
\newblock In \emph{Advances in {Neural} {Information} {Processing} {Systems}},
  volume~27. Curran Associates, Inc., 2014.
\newblock URL \url{https://arxiv.org/abs/1402.1869}.

\bibitem[Nie et~al.(2018)Nie, Zhang, and Patel]{nie_theoretical_2018}
Weili Nie, Yang Zhang, and Ankit Patel.
\newblock A {Theoretical} {Explanation} for {Perplexing} {Behaviors} of
  {Backpropagation}-based {Visualizations}.
\newblock In \emph{Proceedings of the 35th {International} {Conference} on
  {Machine} {Learning}}, pp.\  3809--3818. PMLR, July 2018.
\newblock URL \url{https://proceedings.mlr.press/v80/nie18a.html}.
\newblock ISSN: 2640-3498.

\bibitem[Samek et~al.(2021{\natexlab{a}})Samek, Arras, Osman, Montavon, and
  M{\"u}ller]{samek2021conv_rnn}
Wojciech Samek, Leila Arras, Ahmed Osman, Gr{\'e}goire Montavon, and
  Klaus-Robert M{\"u}ller.
\newblock Explaining the decisions of convolutional and recurrent neural
  networks.
\newblock 2021{\natexlab{a}}.

\bibitem[Samek et~al.(2021{\natexlab{b}})Samek, Montavon, Lapuschkin, Anders,
  and Müller]{samek_explaining_2021}
Wojciech Samek, Grégoire Montavon, Sebastian Lapuschkin, Christopher~J.
  Anders, and Klaus-Robert Müller.
\newblock Explaining {Deep} {Neural} {Networks} and {Beyond}: {A} {Review} of
  {Methods} and {Applications}.
\newblock \emph{Proceedings of the IEEE}, 109\penalty0 (3):\penalty0 247--278,
  March 2021{\natexlab{b}}.
\newblock ISSN 1558-2256.
\newblock \doi{10.1109/JPROC.2021.3060483}.
\newblock Conference Name: Proceedings of the IEEE.

\bibitem[Santoro et~al.(2017)Santoro, Raposo, Barrett, Malinowski, Pascanu,
  Battaglia, and Lillicrap]{santoro2017rn}
Adam Santoro, David Raposo, David~G Barrett, Mateusz Malinowski, Razvan
  Pascanu, Peter Battaglia, and Timothy Lillicrap.
\newblock A simple neural network module for relational reasoning.
\newblock In I.~Guyon, U.~Von Luxburg, S.~Bengio, H.~Wallach, R.~Fergus,
  S.~Vishwanathan, and R.~Garnett (eds.), \emph{Advances in Neural Information
  Processing Systems}, volume~30. Curran Associates, Inc., 2017.
\newblock URL
  \url{https://proceedings.neurips.cc/paper/2017/file/e6acf4b0f69f6f6e60e9a815938aa1ff-Paper.pdf}.

\bibitem[Schulz et~al.(2020)Schulz, Sixt, Tombari, and
  Landgraf]{schulz2020Restricting}
Karl Schulz, Leon Sixt, Federico Tombari, and Tim Landgraf.
\newblock Restricting the flow: Information bottlenecks for attribution.
\newblock In \emph{International Conference on Learning Representations}, 2020.
\newblock URL \url{https://openreview.net/forum?id=S1xWh1rYwB}.

\bibitem[Shah et~al.(2021)Shah, Jain, and Netrapalli]{shah2021grads}
Harshay Shah, Prateek Jain, and Praneeth Netrapalli.
\newblock Do input gradients highlight discriminative features?
\newblock In M.~Ranzato, A.~Beygelzimer, Y.~Dauphin, P.S. Liang, and J.~Wortman
  Vaughan (eds.), \emph{Advances in Neural Information Processing Systems},
  volume~34, pp.\  2046--2059. Curran Associates, Inc., 2021.
\newblock URL
  \url{https://proceedings.neurips.cc/paper/2021/file/0fe6a94848e5c68a54010b61b3e94b0e-Paper.pdf}.

\bibitem[Shapley(1951)]{shapley1951}
Lloyd~S. Shapley.
\newblock \emph{Notes on the N-Person Game – II: The Value of an N-Person
  Game}.
\newblock RAND Corporation, Santa Monica, CA, 1951.
\newblock \doi{10.7249/RM0670}.

\bibitem[Shrikumar et~al.(2016)Shrikumar, Greenside, Shcherbina, and
  Kundaje]{avanti2016blackbox}
Avanti Shrikumar, Peyton Greenside, Anna Shcherbina, and Anshul Kundaje.
\newblock Not just a black box: Learning important features through propagating
  activation differences, 2016.
\newblock URL \url{https://arxiv.org/abs/1605.01713}.

\bibitem[Singla et~al.(2020)Singla, Pollack, Chen, and
  Batmanghelich]{Singla2020Explanation}
Sumedha Singla, Brian Pollack, Junxiang Chen, and Kayhan Batmanghelich.
\newblock Explanation by progressive exaggeration.
\newblock In \emph{International Conference on Learning Representations}, 2020.
\newblock URL \url{https://openreview.net/forum?id=H1xFWgrFPS}.

\bibitem[Sixt et~al.(2020)Sixt, Granz, and Landgraf]{sixt2020wel}
Leon Sixt, Maximilian Granz, and Tim Landgraf.
\newblock When explanations lie: Why many modified {BP} attributions fail.
\newblock In Hal~Daumé III and Aarti Singh (eds.), \emph{Proceedings of the
  37th International Conference on Machine Learning}, volume 119 of
  \emph{Proceedings of Machine Learning Research}, pp.\  9046--9057. PMLR,
  13--18 Jul 2020.
\newblock URL \url{https://proceedings.mlr.press/v119/sixt20a.html}.

\bibitem[Springenberg et~al.(2014)Springenberg, Dosovitskiy, Brox, and
  Riedmiller]{springenberg2014striving}
Jost~Tobias Springenberg, Alexey Dosovitskiy, Thomas Brox, and Martin
  Riedmiller.
\newblock Striving for simplicity: The all convolutional net.
\newblock \emph{arXiv preprint arXiv:1412.6806}, 2014.

\bibitem[{\v{S}}trumbelj \& Kononenko(2011){\v{S}}trumbelj and
  Kononenko]{vstrumbelj2011general}
Erik {\v{S}}trumbelj and Igor Kononenko.
\newblock A general method for visualizing and explaining black-box regression
  models.
\newblock In \emph{International Conference on Adaptive and Natural Computing
  Algorithms}, pp.\  21--30. Springer, 2011.

\bibitem[{\v{S}}trumbelj \& Kononenko(2014){\v{S}}trumbelj and
  Kononenko]{vstrumbelj2014explaining}
Erik {\v{S}}trumbelj and Igor Kononenko.
\newblock Explaining prediction models and individual predictions with feature
  contributions.
\newblock \emph{Knowledge and information systems}, 41\penalty0 (3):\penalty0
  647--665, 2014.

\bibitem[Sundararajan et~al.(2017)Sundararajan, Taly, and
  Yan]{sundararajan2017axiomatic}
Mukund Sundararajan, Ankur Taly, and Qiqi Yan.
\newblock Axiomatic attribution for deep networks.
\newblock In \emph{International conference on machine learning}, pp.\
  3319--3328. PMLR, 2017.

\bibitem[Szegedy et~al.(2013)Szegedy, Zaremba, Sutskever, Bruna, Erhan,
  Goodfellow, and Fergus]{szegedy2013intriguing}
Christian Szegedy, Wojciech Zaremba, Ilya Sutskever, Joan Bruna, Dumitru Erhan,
  Ian Goodfellow, and Rob Fergus.
\newblock Intriguing properties of neural networks.
\newblock \emph{arXiv preprint arXiv:1312.6199}, 2013.

\bibitem[Toms et~al.(2020)Toms, Barnes, and Ebert-Uphoff]{toms2020physically}
Benjamin~A Toms, Elizabeth~A Barnes, and Imme Ebert-Uphoff.
\newblock Physically interpretable neural networks for the geosciences:
  Applications to earth system variability.
\newblock \emph{Journal of Advances in Modeling Earth Systems}, 12\penalty0
  (9):\penalty0 e2019MS002002, 2020.

\bibitem[Viering et~al.(2019)Viering, Wang, Loog, and
  Eisemann]{viering2019manipulate}
Tom Viering, Ziqi Wang, Marco Loog, and Elmar Eisemann.
\newblock How to manipulate cnns to make them lie: the gradcam case.
\newblock \emph{arXiv preprint arXiv:1907.10901}, 2019.

\bibitem[Waeldchen et~al.(2021)Waeldchen, Macdonald, Hauch, and
  Kutyniok]{waeldchen2021computational}
Stephan Waeldchen, Jan Macdonald, Sascha Hauch, and Gitta Kutyniok.
\newblock The computational complexity of understanding binary classifier
  decisions.
\newblock \emph{J. Artif. Int. Res.}, 70:\penalty0 351–387, may 2021.
\newblock ISSN 1076-9757.
\newblock \doi{10.1613/jair.1.12359}.
\newblock URL \url{https://doi.org/10.1613/jair.1.12359}.

\bibitem[Wang et~al.(2020)Wang, Tuyls, Wallace, and
  Singh]{wang-etal-2020-gradient}
Junlin Wang, Jens Tuyls, Eric Wallace, and Sameer Singh.
\newblock Gradient-based analysis of {NLP} models is manipulable.
\newblock In \emph{Findings of the Association for Computational Linguistics:
  EMNLP 2020}, pp.\  247--258, Online, November 2020. Association for
  Computational Linguistics.
\newblock \doi{10.18653/v1/2020.findings-emnlp.24}.
\newblock URL \url{https://aclanthology.org/2020.findings-emnlp.24}.

\bibitem[Xiong et~al.(2020)Xiong, Huang, Yu, Liu, Zhu, and
  Shao]{xiong_number_2020}
H.~Xiong, L.~Huang, M.~Yu, L.~Liu, F.~Zhu, and L.~Shao.
\newblock On the {Number} of {Linear} {Regions} of {Convolutional} {Neural}
  {Networks}.
\newblock In \emph{{ICML}}, 2020.

\bibitem[Zhang et~al.(2021)Zhang, Madumal, Miller, Ehinger, and
  Rubinstein]{zhang2021invertible}
Ruihan Zhang, Prashan Madumal, Tim Miller, Krista~A Ehinger, and Benjamin~IP
  Rubinstein.
\newblock Invertible concept-based explanations for cnn models with
  non-negative concept activation vectors.
\newblock In \emph{Proceedings of the AAAI Conference on Artificial
  Intelligence}, volume~35, pp.\  11682--11690, 2021.

\end{thebibliography}
\bibliographystyle{tmlr}

\newpage

\appendix

\section{Proofs}
\label{appendix:proofs}

\subsection{Proof of proposition \ref{prop:recursive_taylor_relu}
and \ref{prop:recursive_taylor_equals_gradient}
}
\label{appendix:proofs_recursive_taylor}

We will proof Propositions
\ref{prop:recursive_taylor_equals_gradient},
\ref{prop:recursive_taylor_relu}, and
\ref{prop:higher_order_terms} together.

We start with the partial derivative of the relevance function at layer $l$:
\begin{align}
\frac{\partial R_\idx{k}^{l}(\va_l)}{\partial \va_l} &=
\sum_{j = 1}^{d_{l+1}}
    \frac{\partial}{\partial \va_l}
    \left(
        \left[
            \frac
                {
                    \partial R^{l+1}_\idx{j}\left(
                        f_l(\va_{l})
                    \right)
                }
                {\partial \va_{l_\idx{k}}}
        \right]_{\va_{l} =\tilde \va_l^{(j)}(\va_{l})}
        \cdot
        \left\{\va_l - \tilde \va_{l}^{(j)}(\va_{l})\right\}_\idx{k}
    \right)
 \\
& =
\sum_{j = 1}^{d_{l+1}}
    \left(
        \frac
            { \partial \left(\va_{l_\idx{k}} - \tilde \va_{l_\idx{k}}^{(j)}(\va_{l})\right) }
            {\partial \va_{l}}
        \cdot
        \left[
        \frac
            {
                \partial R^{l+1}_\idx{j}\left(
                    f_l(\va_{l})
                \right)
            }
            {\partial \va_{l}}
        \right]_{\va_{l} =\tilde \va_l^{(j)}(\va_{l})}
    \right.
    \\
    &
    \left.
    \quad \quad \quad \quad +
        \underbrace{
            \frac
            {\partial}
            {\partial \va_l}
            \cdot
            \left[
            \frac
                {
                    \partial R^{l+1}_\idx{j}\left(
                        f_l(\va_{l})
                    \right)
                }
                {\partial \va_{l}}
            \right]_{\va_{l} =\tilde \va_l^{(j)}(\va_{l})}
            \cdot
            \left\{\va_l - \tilde \va_{l}^{(j)}(\va_{l})\right\}_\idx{k}
        }_{
        =\,0,\text{ for ReLU networks}
        }
    \right)
\end{align}
In this first step, we applied the product rule.
For ReLU networks, the higher-order terms are zero.
For other networks (Transformer, LSTMs), the higher-order terms will not be zero.
The terms which are zero for ReLU networks are exactly the terms from Proposition \ref{prop:higher_order_terms}.

In the next step, we will apply the chain rule and
rewrite $\partial \va_{l_\idx{k}} / \partial \va_{l}$ as the
$k$-standard basis $e_{k}$ (a one-hot vector where the
$k$-th dimension is 1):
\begin{align}
\frac{\partial R_\idx{k}^{l}(\va_l)}{\partial \va_l}
&=
\sum_{j = 1}^{d_{l+1}}
    \left(
        e_{k}
        - \frac
            {\partial \tilde \va_{l_\idx{k}}^{(j)}(\va_{l})}
            {\partial \va_{l}}
    \right)
    \cdot
    \left[
    \frac
        {\partial f_l(\va_{l})}
        {\partial \va_{l}}
    \cdot
    \frac
        {
            \partial R^{l+1}_\idx{j}\left(
                f_l(\va_{l})
            \right)
        }
        {\partial
                f_l(\va_{l})
        }
    \right]_{\va_{l} =\tilde \va_l^{(j)}(\va_{l})},
\end{align}
The next observation is that the gradients inside the $[\ldots]_{\va_{l} =\tilde
\va_l^{(j)}(\va_{l})}$ must be the same for $\va_l$ and the root point $\tilde
\va_l^{(j)}$ as
both are in the same local region of $f_l \circ R_\idx{j}^{l+1}$.
Therefore, we can safely drop the evaluation of the gradient at the root point
($[\ldots]_{\va_{l} =\tilde \va_l^{(j)}(\va_{l})}$) and write:
\begin{align}
\frac{\partial R_\idx{k}^{l}(\va_l)}{\partial \va_l}
&=
\sum_{j = 1}^{d_{l+1}}
    \left(
        \frac
            {\partial f_l(\va_{l})}
            {\partial \va_{l_\idx{k}}}
        - \frac
            {\partial \tilde \va_{l_\idx{k}}^{(j)}(\va_{l})}
            {\partial \va_{l}}
        \cdot
        \frac
            {\partial f_l(\va_{l})}
            {\partial \va_{l}}
    \right)
    \cdot
    \frac
        {
            \partial R^{l+1}_\idx{j}\left(
                f_l(\va_{l})
            \right)
        }
        {\partial
                f_l(\va_{l})
        }
\end{align}
Substituting this result into the definition of $R^{l-1}(\va_{l-1})$
from \eqref{eq:dtd_recursive_taylor_step} yields the result of Proposition \ref{prop:recursive_taylor_relu}.

To proof proposition \ref{prop:recursive_taylor_equals_gradient},
we use $ \partial \tilde \va_{l_\idx{k}}^{(j)}(\va_{l}) / \partial \va_{l} = 0 $ and get:
\begingroup
\allowdisplaybreaks
\begin{align}
\frac{\partial R_\idx{k}^{l}(\va_l)}{\partial \va_l}
&=
\sum_{j = 1}^{d_{l+1}}
    \frac
            {\partial f_l(\va_{l})}
            {\partial \va_{l_\idx{k}}}
    \cdot
    \frac
        {
            \partial R^{l+1}_\idx{j}\left(
                f_l(\va_{l})
            \right)
        }
        {\partial
                f_l(\va_{l})
        }
\\
&=
    \frac
            {\partial f_l(\va_{l})}
            {\partial \va_{l_\idx{k}}}
    \cdot
    \sum_{j = 1}^{d_{l+1}}
    \frac
        {
            \partial R^{l+1}_\idx{j}\left(
                f_l(\va_{l})
            \right)
        }
        {\partial
                f_l(\va_{l})
        }
\\
& =
    \frac
        { \partial f_l(\va_{l}) }
        {\partial \va_{l_\idx{k}}}
    \cdot
    \sum_{j = 1}^{d_{l+1}}
        \frac
            {\partial f_{l+1}(\va_{l+1}) }
            {\partial \va_{{l+1}_\idx{j}} }
        \cdot
        \sum_{i = 1}^{d_{l+2}}
        \frac
            {
                \partial R^{l+2}_\idx{i}\left(
                    f_{l+1}(\va_{l+1})
                \right)
            }
            {\partial f_{l+1}(\va_{l+1}) }
\\
& =
    \frac
        { \partial f_l(\va_{l}) }
        {\partial \va_{l_\idx{k}}}
    \cdot
    \frac
        {\partial f_{l+1}(\va_{l+1}) }
        {\partial \va_{{l+1}} }
    \cdot
    \sum_{i = 1}^{d_{l+2}}
    \frac
            {
                \partial R^{l+2}_\idx{i}\left(
                    f_{l+1}(\va_{l+1})
                \right)
            }
            {\partial f_{l+1}(\va_{l+1}) }
\\
& =
    \frac
        { \partial f_l(\va_{l}) }
        {\partial \va_{l_\idx{k}}}
    \cdot
    \frac
        {\partial f_{l+1}(\va_{l+1}) }
        {\partial \va_{{l+1}} }
    \cdot
    \ldots
    \cdot
    \frac
        {\partial f_{n-1}^{n}(\va_{n-1}) }
        {\partial \va_{{n-1}} }
    \cdot
    \sum_{i = 1}^{d_{n+1}}
    \frac
            {
                \partial R^{n+1}_\idx{i}\left(
                    f_{n}^{n+1}(\va_{n})
                \right)
            }
            {\partial f_{n}^{n+1}(\va_{n}) }
\\
& =
    \frac
        { \partial f_l(\va_{l}) }
        {\partial \va_{l_\idx{k}}}
    \cdot
    \frac
        {\partial f_{l+1}(\va_{l+1}) }
        {\partial \va_{{l+1}} }
    \cdot
    \ldots
    \cdot
    \frac
        {\partial f_{n-1}(\va_{n-1}) }
        {\partial \va_{{n-1}} }
    \cdot
    \frac
        {\partial f_{n}(\va_{n}) }
        {\partial \va_{{n}} }
    \cdot
    e_\xi
\\
& =
\nabla f_{l_\idx{\xi}}(\va_l)
\end{align}

Substituting this into the relevance function of the input $R(\vx) = R^1(\va_1)$ and
using $\frac{\partial R(\vx)}{\partial \vx}\big|_{\vx = \tilde \vx(\vx)} =
\frac{\partial R(\vx)}{\partial \vx}$ (as $\tilde \vx$ must be in the same linear region),
yields:
\begin{equation}
    R(\vx) = R(\tilde \vx(\vx)) +
        \left.
            \frac{\partial R(\vx)}{\partial \vx}
        \right|_{\vx = \tilde \vx(\vx)}
        \odot
        (\vx - \tilde \vx)
        = R(\tilde \vx(\vx)) + \nabla f_{\idx{\xi}}(\vx)
        \odot
        (\vx - \tilde \vx(\vx)),
\end{equation}
which finished the proof of Proposition \ref{prop:recursive_taylor_equals_gradient}.
\endgroup

\subsection{Admissible Region for the root points of the Relevance Function}
\label{appendix:admissible_region_proof}

We now proof Proposition \ref{prop:admissible_regions_root_points} which is restated here:

\textbf{Proposition \ref{prop:admissible_regions_root_points}}
(C1: Recursive Taylor cannot increase the size of admissible regions)
\emph{
Given a ReLU network $f: \R^{d_1} \to \R_{\ge 0}^{d_{n+1}}$, recursive
relevance functions $R^l(\va_l)$ with $l \in \{1,\ldots,n\}$ according to definition
\ref{def:dtd_recursive_taylor}, and let $\xi$ index the explained logit.
Then it holds for the admissible region $N_{R^l}(\va_l)$ for the root points $\tilde \va_l^{(j)}$
of the relevance function $R^l$ that
$N_{R^l}(\va_l) \subseteq N_{f_{n_\xi}\circ \ldots \circ f_{l}}(\va_l)$.
}

Let $\tilde \va_1, \ldots, \tilde \va_n$ fix the root points.
Proof by induction over the number of layers. We start with the induction base case at the
final layer.  There, we have $R^n(\tilde \va_n) = f_\idx{\xi}(\tilde \va_n)$, which follows
from the recursion base case.
Clearly, $N_{R^n}(\tilde \va_n) \subseteq N_{f_{n_\xi}}(\tilde \va_n)$.
Induction step:
We assume $N_{R_{l+1}}(\tilde \va_{l+1}) \subseteq N_{f_{l+1}}(\tilde \va_{l+1})$.
For the layer $l$, the root points must be valid for the function
$R^{l+1}(f_l(\tilde \va_l))$. As we know that
$N_{R_{l+1}}(\tilde \va_{l+1}) \subseteq N_{f_{l+1}}(\tilde \va_{l+1})$,
it must also be the case that $N_{R^l}(\tilde \va_l) \subseteq N_{f_l}(\tilde \va_l)$.

\section{Details About the Relation Network for the CLEVR dataset}
\label{app:clevr_model}

Our code-base builds upon a public available implementation of relation
networks\footnote{\url{https://github.com/rosinality/relation-networks-pytorch}}
and utilized Captum for computing \LRPa explanations \cite{kokhlikyan2020captum}.
We also setup the CLEVR XAI dataset released on
Github\footnote{\url{https://github.com/ahmedmagdiosman/clevr-xai/releases/tag/v1.0}}

\section{Pseudo-Code}

\subsection{Full-backward DTD}
\label{sec:pseudo_code_fullbackward_dtd}
\begin{algorithm}[H]
    \caption{
        Pseudocode for the recursive application of the Taylor Theorem.
    The  global state contains the following variables:
     $f_1, \ldots, f_{n}$ the layer functions of the network,
     $d_1, \ldots, d_{n+1}$ the dimension of the input to each layer,
     and $\xi$ the index of the output neuron.
    }
    \label{alg:dtd_recursive_taylor}
    \begin{algorithmic}
    \Statex
    \Function{get\_relevance}{$l$: layer index, $\va_l$: the layer input}
        \If{$l = n + 1$}
            \State \Return $\va_{l_\idx{\xi}}$
        \EndIf
        \State $\tilde \va_l \gets $ \Call{find\_root\_point}{$f, l, \va_l$}
        \State $R^{l+1} \gets $ \Call{get\_relevance}{$l+1, f_l(\tilde \va_l)$}
        \For{$j \in 1\ldots d_{l+1}$}
            \State $\tilde \va\method{grad} \gets 0$
            \State $R_\idx{j}^{l+1}\method{backward}()$
            \State $\vr_j \gets \tilde \va\method{grad}\odot (\va_l - \tilde \va_{l})$
        \EndFor
        \State \Return $\sum_{j = 1} ^{d_{l+1}} \vr_j$
    \EndFunction
    \end{algorithmic}
\end{algorithm}

\newpage

\subsection{Train-free DTD}
\label{sec:pseudo_code_train_free_dtd}
\begin{algorithm}[h]
    \caption{DTD Train-Free}
    \label{alg:dtd_train_free}
    \begin{algorithmic}
    \Statex
    \Function{find\_root\_point}
        {$l$: layer index, $\va_l$: the layer input, $R^{l+1}_\idx{j}$: the relevance}
        \State $\vw_j = W_\idx{j:}$
        \If { $z^+$-rule}
            \State $\vv = \va_l \odot \ind_{\vw_j\ge0} $
        \ElsIf { $w^2$-rule}
            \State $\vv = \vw_j $
        \ElsIf { $\gamma$-rule}
            \State $\vv = \va_l( 1 + \gamma \ind_{\vw_j\ge0}) $
        \State \ldots
        \EndIf
        \State $t = R^{l+1}_\idx{j} / (\vw_j^T \vv)$
        \State \Return $\va_l - t \vv$ \Comment{
            Ensures that $\vw_j^T (\va_l - \tilde \va_l)
                = \vw_j^T\frac{R_\idx{j}}{\vw_j^T \vv} \vv
                = R_\idx{j}$
        }
    \EndFunction
    \Statex
    \Function{get\_relevance}{$l$: layer index, $\va_l$: the layer input}
        \If{$l = n + 1$}
            \State \Return $\va_{l_\idx{\xi}}$
        \EndIf
        \State $R^{l+1} \gets $ \Call{get\_relevance}{$l+1, f_l(\va_l)$}
        \Comment{relevance of input instead of root point $\tilde \va_l$}
        \For{$j \in 1\ldots d_{l+1}$}
            \State $\tilde \va_l^{(j)} \gets $ \Call{find\_root\_point}{$f, l, \va_l, R^{l+1}_\idx{j}$}
            \State $\tilde \va\method{grad} \gets 0$
            \State $\vo \gets [W_{l_\idx{j}} \va_l^{(j)} + \vb_{l_\idx{j}}]$
            \State $\vo$\method{backward()}
            \State $\vr_j \gets \tilde \va\method{grad}\odot (\va_l - \tilde \va_{l})$
        \EndFor
        \State \Return $\sum_{j = 1} ^{d_{l+1}} \vr_j$
    \EndFunction
    \end{algorithmic}
\end{algorithm}

\end{document}